\documentclass[acmsmall]{acmart}

\usepackage{url}
\usepackage{booktabs} 
\usepackage{algorithmic}
\floatname{algorithm}{Procedure}
\usepackage{colortbl}
\usepackage[ruled]{algorithm2e} 

\usepackage{subcaption}
\usepackage{multirow}
\usepackage{tabularx}
\usepackage{bm}
\usepackage{graphicx}
\usepackage{rotating}
\usepackage{mathrsfs}

\def\bf{\textbf}
\acmJournal{JACM}
\acmVolume{37}
\acmNumber{4}
\acmArticle{111}
\acmMonth{8}

\AtBeginDocument{%
  \providecommand\BibTeX{{%
    \normalfont B\kern-0.5em{\scshape i\kern-0.25em b}\kern-0.8em\TeX}}}

\setcopyright{acmcopyright}
\copyrightyear{2018}
\acmYear{2018}
\acmDOI{10.1145/1122445.1122456}

\begin{document}

\title[DeepKey: A Multimodal Biometric Authentication System...]{DeepKey: A Multimodal Biometric Authentication System via Deep Decoding Gaits and Brainwaves} 
 

\author{Xiang Zhang}
\affiliation{%
  \institution{University of New South Wales}
  \streetaddress{CSE, UNSW}
  \city{Sydney}
  \state{NSW}
  \postcode{2052}
  \country{Australia}}
  \email{xiang.zhang3@student.unsw.edu.au}

\author{Lina Yao}
\affiliation{%
  \institution{University of New South Wales}
  \streetaddress{CSE, UNSW}
  \city{Sydney}
  \state{NSW}
  \postcode{2052}
  \country{Australia}}
  \email{lina.yao@unsw.edu.au}

\author{Chaoran Huang}  
\affiliation{%
  \institution{University of New South Wales}
  \streetaddress{CSE, UNSW}
  \city{Sydney}
  \state{NSW}
  \postcode{2052}
  \country{Australia}}
  \email{chaoran.huang@unsw.edu.au}

\author{Tao Gu} 
\affiliation{%
  \institution{RMIT University}
  \streetaddress{}
  \city{Melbourne}
  \state{VIC}
  \postcode{3001}
  \country{Australia}}
  \email{tao.gu@rmit.edu.au}

 \author{Zheng Yang}
\affiliation{%
  \institution{Tsinghua University}
  \streetaddress{School of Software, Tsinghua University}
  \city{Beijing}
  \state{Beijing}
  \postcode{100084}
  \country{China}}
  \email{hmilyyz@gmail.com}

 \author{Yunhao Liu}
\affiliation{%
  \institution{Michigan State University}
  \streetaddress{}
  \city{East Lansing}
  \state{Michigan}
  \postcode{48824}
  \country{USA}}
  \email{yunhao@cse.msu.edu}

\renewcommand{\shortauthors}{Xiang Zhang, et al.}

\begin{abstract}
Biometric authentication involves various technologies to identify individuals by exploiting their unique, measurable physiological and behavioral characteristics. However, traditional biometric authentication systems (e.g., face recognition, iris, retina, voice, and fingerprint) are facing an increasing risk of being tricked by biometric tools such as anti-surveillance masks, contact lenses, vocoder, or fingerprint films. In this paper, we design a multimodal biometric authentication system named Deepkey, which uses both Electroencephalography (EEG) and gait signals to better protect against such risk. Deepkey consists of two key components: an Invalid ID Filter Model to block unauthorized subjects and an identification model based on attention-based Recurrent Neural Network (RNN) to identify a subject's EEG IDs and gait IDs in parallel. 
The subject can only be granted access while all the components produce the consistent evidence to match the user's proclaimed identity. We implement Deepkey with a live deployment in our university and conduct extensive empirical experiments to study its technical feasibility in practice. DeepKey achieves the False Acceptance Rate (FAR) and the False Rejection Rate (FRR) of \textbf{0} and \textbf{1.0\%}, respectively.
The preliminary results demonstrate that Deepkey is feasible, show consistent superior performance compared to a set of methods, and has the potential to be applied to the authentication deployment in real-world settings.
\end{abstract}


\maketitle

\section{Introduction} 
\label{sec:Introduction}
Over the past decades, biometric authentication systems have gained popularity due to the reliability and adaptability. Existing biometric authentication systems generally include physiological and behavioral ones. The former is based on individuals' unique intrinsic features (e.g., face \cite{givens2013biometric}, iris \cite{latman2013field}, retina \cite{sadikoglu2016biometric}, voice \cite{goldstein2016methods}, and fingerprint \cite{unar2014review}) and the latter is based on individuals' behavior patterns such as gait analysis \cite{boulgouris2013gait}.
Recently, biometrics (e.g., fingerprint, face) based authentication systems face an increasing threat of being deceived as a result of the rapid development of manufacturing industry and technologies. For example, individuals can easily trick a fingerprint-based authentication system by using a fake fingerprint film\footnote{\url{http://www.instructables.com/id/How-To-Fool-a-Fingerprint-Security\-System-As-Easy-/}} 
or an expensive face recognition-based authentication systems by simply wearing a two-hundred-dollar 
anti-surveillance mask\footnote{\url{http://www.urmesurveillance.com/urme-prosthetic/}}. 
Thus, fake-resistance characteristics are becoming a more significant requirement for any authentication system. 
To address the aforementioned issues, EEG (Electroencephalography) signal-based cognitive biometrics and gait-based systems have been attracting increasing attention. 

\begin{table}[t]
\centering
\caption{Comparison of various biometrics. EEG and Gait have considerable fake-resistance which is the most significant characteristic of authentication systems. $\uparrow$ denotes the higher the better while $\downarrow$ denotes the lower the better.}
\label{tab:comparison_1}
\resizebox{\textwidth}{!}{\begin{tabular}{lllllllll}
\hline 
 & \textbf{Biometrics} & \begin{tabular}[c]{@{}l@{}}\bf{Fake-}\\ \bf{-resistance}\end{tabular} $\uparrow$ & \textbf{Universality} $\uparrow$ & \textbf{Uniqueness} $\uparrow$ & \textbf{Stability} $\uparrow$ & \textbf{Accessibility} $\uparrow$ & \textbf{Performance} $\uparrow$ & \begin{tabular}[c]{@{}l@{}}\bf{Computational}\\ \bf{Cost}\end{tabular} $\downarrow$ \\ \hline
\multirow{8}{*}{\begin{tabular}[c]{@{}l@{}}\textbf{Uni-}\\ \textbf{-modal}\end{tabular}} & Face/Vedio & Medium & Medium & Low & Low & High & Low & High \\
 & Fingerprint/Palmprint & Low & High & High & High & Medium & High & Medium \\
 & Iris & Medium & High & High & High & Medium & High & High \\
 & Retina & High & Medium & High & Medium & Low & High & High \\
 & Signature & Low & High & Low & Low & High & Low & Medium \\
 & Voice & Low & Medium & Low & Low & Medium & Low & Low \\
 & Gait & High & Medium & High & Medium & Medium & High & Low \\
 & EEG & High & Low & High & Low & Low & High & Low \\ \hline
\multirow{6}{*}{\begin{tabular}[c]{@{}l@{}}\textbf{Mulit-}\\ \textbf{-modal}\end{tabular}} & Fingerprint+face & Low & High & Low & Low & High & Medium & High \\
 & Iris+pupil & Medium & Medium & High & High & Medium & High & High \\
 & Iris+face & Medium & High & Medium & Medium & Medium & Medium & High \\
 & ECG+fingerprint & High & Medium & Medium & High & Low & High & Medium \\
 \hline
 & \bf{ EEG+gait}& \bf{ High} & \bf{ Low} & \bf{ High} & \bf{ High} & \bf{ Low} & \bf{ High} & \bf{ Low} \\ \hline
\end{tabular}
}
\end{table}

EEG signal-based authentication systems are an emerging approach in physiological biometrics. 
EEG signals measure the brain's response and record the electromagnetic, invisible, and untouchable electrical neural oscillations. Many research efforts have been made on EEG-based biometric authentication for the uniqueness and reliability .
EEG data are unique for each person and almost impossible to be cloned and duplicated. Therefore, an EEG-based authentication system has the potential to uniquely identify humans and ingenious enough to protect against faked identities~\cite{chuang2013think}. 
For instance, Chuang et al. \cite{chuang2013think}propose a single-channel EEG-based authentication system, which achieves an accuracy of 0.99. Sarineh Keshishzadeh et al. \cite{keshishzadeh2016improved} employ a statistical model for analyzing EEG signals and achieves an accuracy of 0.974. Generally, EEG signals have the following inherent advantages:
\begin{itemize}
  \item \textit{Fake-resistibility.} EEG data are unique for each person and almost impossible to be cloned and duplicated. 
    EEG signals are individual-dependent. Therefore, an EEG-based authentication system has the potential to verify human identity and ingenious enough to protect against faked identities\cite{chuang2013think}.
  \item \textit{Reliability.} An EEG-based authentication system can prevent the subjects under abnormal situations (e.g., dramatically spiritual fluctuating, hysterical, drunk, or under threaten) since EEG signals are sensitive to human stress and mood.
  \item \textit{Feasibility.} We have seen an important trend to build authentication systems based on EEG because the equipment for collecting EEG data is cheap and easy to acquire, and it is expected to be more precise, accessible, and economical in the future. 
\end{itemize}

In comparison, gait-based authentication systems have been an active direction for years \cite{yao2018compressive,qian2018enabling}. Gait data are more generic and can be gathered easily from popular inertial sensors. Gait data are also unique because they are determined by intrinsic factors (e.g., gender, height, and limb length), 
temporal factors \cite{callisaya2009population} (e.g., step length, walking speed, and cycle time) and kinematic factors (e.g., joint rotation of the hip, knee, and ankle, mean joint angles of the hip/knee/ankle, and thigh/trunk/foot angles). In addition, a person's gait behavior is established inherently in the long-term and therefore difficult to be faked.
Hoang et al. \cite{hoang2014secure} propose a gait-based authentication biometric system to analyze gait data gathered by mobile devices, adopt error correcting codes to process the variation in gait measurement, and finally achieve a false acceptance rate (FAR) of 3.92\% and a false rejection rate (FRR) of 11.76\%. 
Cola et al. \cite{cola2016gait} collect wrist signals and train gait patterns to detect invalid subjects (unauthenticated people). The proposed method achieves an equal error rate (EER) of 2.9\%.

Despite the tremendous efforts,
various other challenges still remain in single EEG/gait based authentication systems:
 (i) the solo EEG/gait authentication system generally obtains a False Acceptance Rate (FAR, which is extremely crucial in high-confidential authentication scenarios)
 higher than $3\%$. It is not precise enough for highly confidential places such as military bases, the treasuries of banks and political offices where tiny misjudges could provoke great economic or political catastrophes; (ii) the single authentication system may break down under attack but no backup plan is provided; (iii) the solo EEG-based authentication system is easy to be corrupted by environmental factors (e.g., noise) and subjective factors (e.g., mental state) although it has high rake-resistance; (iv) the solo gait-based authentication system has relatively low performance although it is more stable over different scenarios.

In this paper, we propose Deepkey, a novel biometric authentication system that enables dual-authentication leveraging on the advantages of both gait-based and EEG-based systems. Compared with either gait-based or EEG-based authentication system, a dual-authentication system offers more reliable and precise identification. 
Table~\ref{tab:comparison_1} summarizes the overall comparison of Deepkey with some representative works on seven key aspects.  
Deepkey consists of three main components: the Invalid ID Filter Model to eliminate invalid subjects, the EEG Identification Model to identify EEG IDs, and the Gait Identification Model to identify gait IDs. 
An individual is granted access only after she/he passes all the authentication components. Our main contributions 
are highlighted as follows:

\begin{itemize}
  \item We present Deepkey, a dual-authentication system that exploits both EEG and gait biological traits. To the best of our knowledge, Deepkey is the first two-factor authentication system for person authentication 
  using EEG and gaits.
  Deepkey is empowered high-level fake-resistance and reliability because both EEG and gait signals are invisible and 
  hard to be reproduced.

  \item We design a robust framework that includes an attention-based RNN to detect and classify of multimodal sensor data, and to decode the large diversity in how people perform gaits and brain activities simultaneously. The delta band of EEG data is decomposed for the rich discriminative information. 
  \item We validate and evaluate Deepkey on several locally collected datasets. The results show that Deepkey significantly outperforms a series of baseline models and the state-of-the-art methods, achieving FAR of 0 and FRR of $1\%$. Further, we design extensive experiments to investigate the impact of key elements.
\end{itemize}
The remainder of this paper is organized as follows. Section~\ref{sec:Related Work} introduces the EEG-based, gait-based, and
multimodal biometric systems briefly. Section~\ref{sec:The Proposed approach} present the methodology framework and three key models (Invalid ID Filter Model, Gait Identification Model, and EEG Identification Model) of the DeepKey authentication system in detail. Section~\ref{sec:experiment_and_results} evaluates the proposed approach on the public Gait and EEG dataset and provides analysis of the experimental results. Finally, Section~\ref{sec:discussions_and_future_work} discusses the opening challenges of this work and highlight the future scope while Section~\ref{sec:conclusion} summarizes the key points this paper.

\section{Related Work} 
\label{sec:Related Work}
In this section, we introduce the related studies on several topics: biometric authentication technologies,  EEG-based authentication, gait-based authentication, and multimodal biometric authentication.

\subsection{Biometric authentication technologies} 
\label{sec:biometric_authentication_technologies}

Since biometric features cannot be stolen or duplicated easily, biometric authentication is becoming increasingly a commonplace. Currently, the most mature biometric authentication technology is fingerprint-based authentication which has been demonstrated to high matching accuracy and been used for decades \cite{maio2002fvc2002}. Iris recognition is another popular approach for biometric authentication owing to its unique and stable pattern \cite{pillai2014cross}. 
In 1993, Daugman \cite{daugman1993high} proposes to use Gabor phase information and Hamming distance for iris code matching, which still is the most classic iris recognition method. Based on \cite{daugman1993high}, a flurry of research \cite{pillai2014cross} has emerged offering solutions to ameliorate iris authentication problems. 
For example, Pillai et al. \cite{pillai2014cross} introduce kernel functions to represent transformations of iris biometrics. This method restrains both the intra-class and inter-class variability to solve the sensor mismatch problem. Face recognition techniques \cite{guzman2013thermal, drira20133d, zhou2014recent} is the most commonly used and accepted by the public for its unique features and non-intrusiveness. Since face recognition systems require tackling different challenges including expression, image quality, illumination, and disguise to achieve high accuracy, infrared (IR) \cite{guzman2013thermal} and 3D \cite{drira20133d} systems have attracted much attention. According to \cite{zhou2014recent}, multimodal recognition combining traditional visual textual features and IR or 3D systems can achieve higher accuracy than single modal systems. \par

\subsection{EEG-based Authentication} 
\label{sec:eeg_based_authentication}

Since EEG can be gathered in a safe and non-intrusive way, researchers have paid great attention to exploring this kind of brain signals. For person authentication, EEG is, on the one hand, promising for being confidential and fake-resistant but, on the other hand, complex and hard to be analyzed. Marcel and Mill{\'a}n \cite{marcel2007person} use Gaussian Mixture Models and train client models with maximum A Posteriori (MAP). Ashby et al. \cite{ashby2011low} extract five sets of features from EEG electrodes and inter-hemispheric data, combine them together, and process the final features with support vector machine (SVM). The study shows that EEG authentication is also feasible with less-expensive devices. Altahat et al. \cite{altahat2015analysing} select Power Spectral Density (PSD) as the feature instead of the widely used autoregressive (AR) models to achieve higher accuracy. They also conduct channel selection to determine the contributing channels among all 64 channels. Thomas and Vinod \cite{thomas2016utilizing} take advantage of individual alpha frequency (IAF) and delta band signals to compose specific feature vector. They also prefer PSD features but only perform the extraction merely on gamma band.


\subsection{Gait-based Authentication} 
\label{sec:gait_based_authentication}

As the most basic activity in our daily lives, walking is an advanced research hotspot for activity recognition \cite{yao2016learning, huang2016exploiting, mennicken2016s}. Differing from previous work, our work focuses on human gait, a spatio-temporal biometric that measures a person's manner 
on walking. On the one hand, gait can be collected remotely without human interaction 
compared 
to 
other aforementioned biometric features \cite{wang2012human}. 
On the other hand, it is challenging to eliminate the influence of exterior factors including clothing, walking surface, shoes, carrying stuff, and other environmental factors. Existing gait recognition approaches sit in two categories. One is {\em model-based approach} \cite{nixon2009model}, which models gait information with mathematical structures, and the other is 
{\em appearance-based approach}, which extracts features in a straightforward way irrespective of the mathematical structure. 
Due to its high efficiency and remarkable performance, Gait Energy Image (GEI) \cite{man2006individual} has become one of the most popular appearance-based methods in recent years. Based on GEIs, 
conducts enhancing process and autocorrelation on the contour profile. Besides, the cross-view variance is also a concern of gait identification \cite{kusakunniran2013new, kusakunniran2014recognizing}. For example, Wu et al. \cite{wu2017comprehensive} consider not only the cross-view variance but also deep convolutional neural networks (CNNs) for robust gait identification. 

\subsection{Multimodal Biometric Authentication} 
\label{sec:multimodal_biometric_authentication}

Since traditional unimodal authentication suffers from the negative influence of loud noise, low universality, and intra-class variation, it cannot achieve higher accuracy in a wide range of applications. To address this issue, multimodal biometric authentication 
which combines and uses biometric traits in different ways, is becoming popular. Taking the commonness into consideration, most works choose two biometrics from face, iris, and fingerprints and make the fusion \cite{telgad2014combination, khiari2016quality, kumar2013feature}. In \cite{derawi2014fusion}, an innovative combination between gait and electrocardiogram (ECG) is shown to be effective. Manjunathswamy et al. \cite{manjunathswamy2015multimodal} combine ECG and fingerprint at the score level. 
To the best of our knowledge, the approach proposed in this paper is the first 
to combine EEG and gaits for person authentication. Taking advantages of both EEG and gait signals, the combination is expected to improve the reliability of authentication systems.

This dual-authentication system is partially based on our previous work MindID \cite{zhang2018mindid} which is an EEG-based identification system. We emphasize several difference compared to \cite{zhang2018mindid}: 1) this work is an authentication system with an invalid ID filter while \cite{zhang2018mindid} only focuses on identification; 2) this work adopt two biometrics including EEG and gait while \cite{zhang2018mindid} only exploit EEG signals; 3) this work conduct extensive real world experiments to collect gait signals.

\section{Deepkey Authentication System} 
\label{sec:The Proposed approach}
In this section, we first report the workflow of Deepkey
 to give
  an overview of the authentication algorithm, and then present the technical details for each component. 

\subsection{Deepkey System Overview} 
\label{sec:system_overview}
\begin{figure}
  \includegraphics[width=\linewidth]{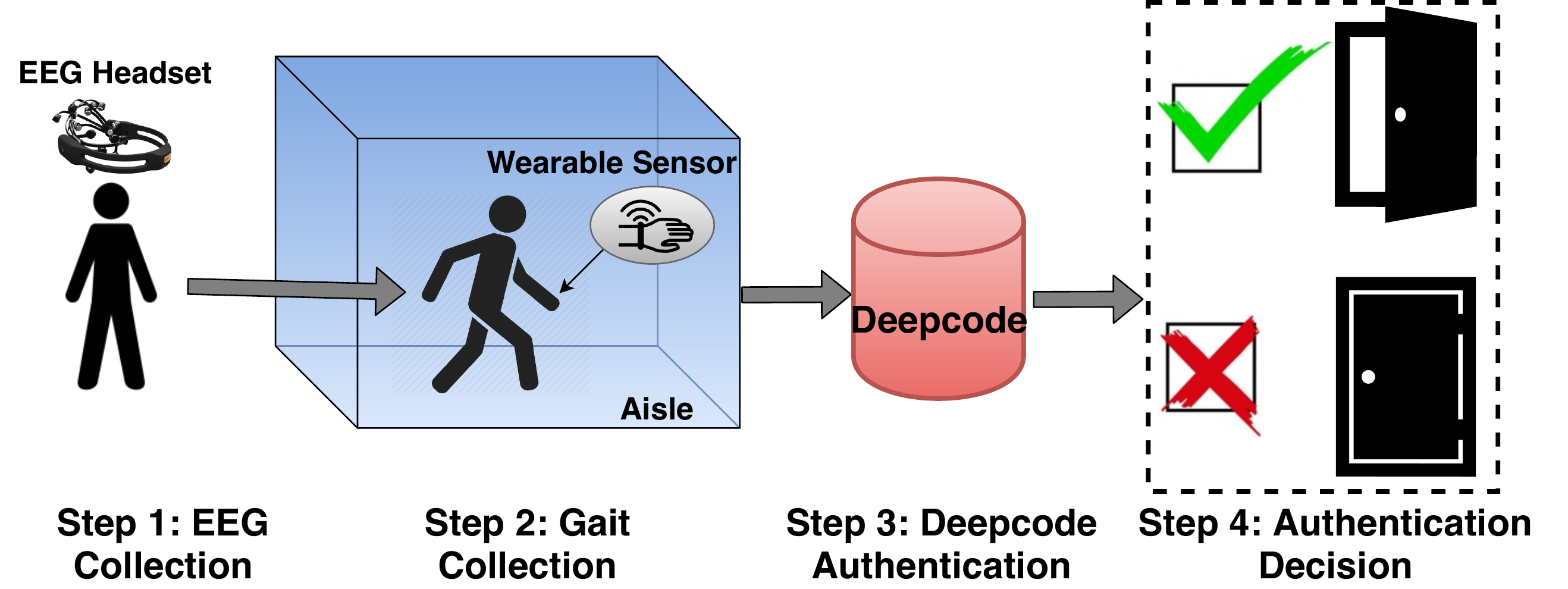}
  \caption{Workflow of Deepkey authentication system. The data collection of EEG and gait are cascade.}
  \label{fig:scenario}
\end{figure}


The Deepkey system is supposed to be deployed in access to confidential locations (e.g., bank vouchers, military bases, and government confidential residences). As shown in Figure~\ref{fig:scenario}, the overall workflow of the Deepkey authentication system consists of the following four steps:
\begin{enumerate}
 \item Step 1: EEG data collection. The subject, who 
  requests for authentication, is required to wear the EEG headset and stays in a relaxed state. 
  The EEG data ($\mathcal{E}$) will typically take 2 seconds.
  \item Step 2: Gait data collection. The subject takes off the EEG headset and puts three IMUs (Inertial Measurement Unit) and walks through an aisle to collect gait data $\mathcal{G}$ by IMUs. 
  \item Step 3: Authentication. The gathered EEG and gait data are flatten and associated with input data $\mathcal{I} =[\mathcal{E}:\mathcal{G}]$ to be fed into the Deepkey authentication algorithm.
  \item Step 4: Decision. An \textit{Approve} or \textit{Deny} decision will be made according to the Deepkey authentication results.
\end{enumerate}

\begin{figure}[t]
  \includegraphics[width=0.8\linewidth]{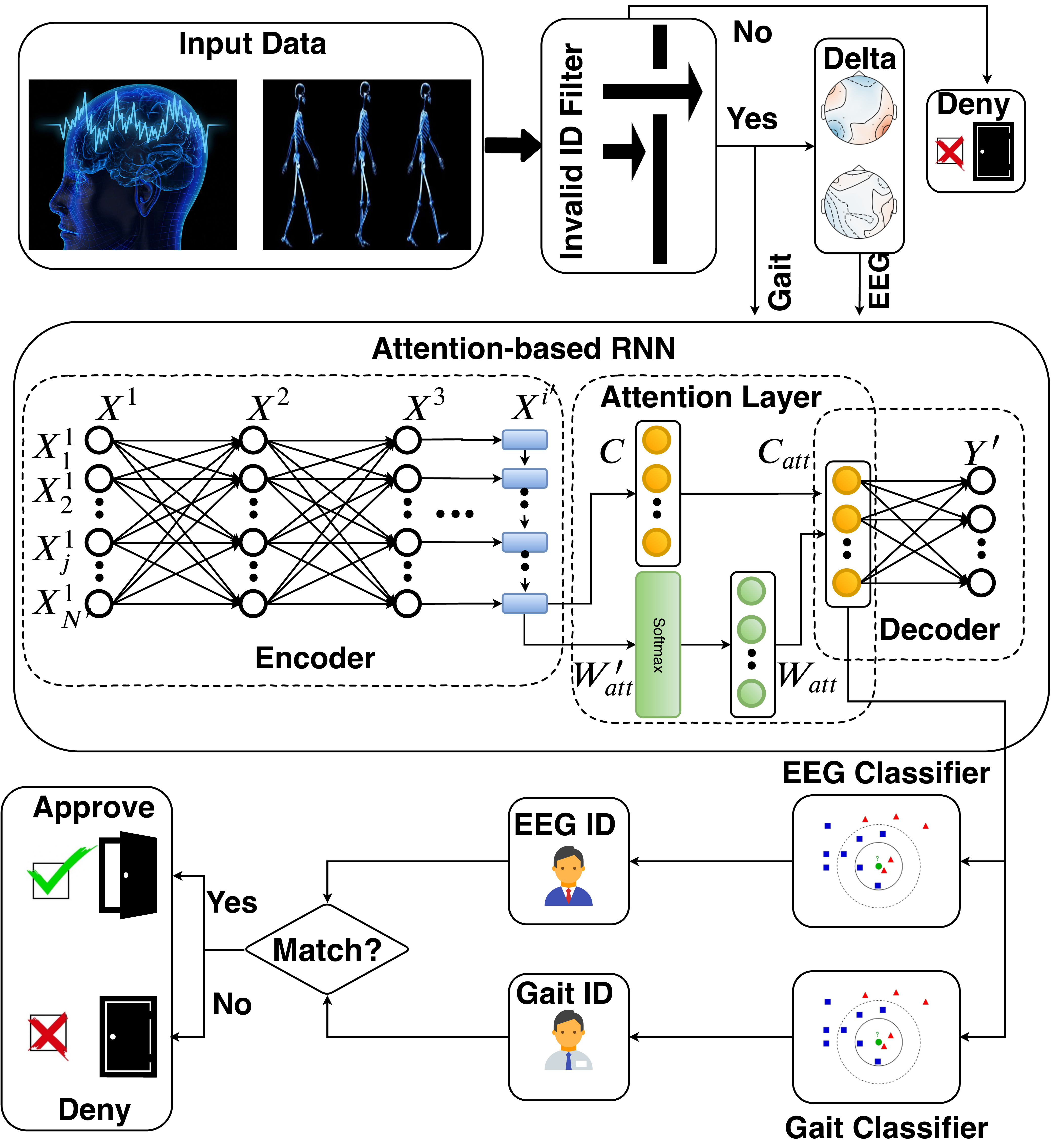}
  \caption{Authentication workflow. If the input data can not pass the invalid ID filter, it would directly regarded as an impostor and deny access. If pass, the Delta pattern and gait signals are parallelly fed into an attention-based RNN structure to study the distinctive features $C_{att}$. The learned features are classified by the EEG and Gait classifier in order to identify the subject's EEG and Gait ID. The subject is approved only if the EEG ID is match with Gait ID.}
  \label{fig:attention}
\end{figure}

The most crucial component among above the steps is the third step, where the Deepkey authentication system receives the associated input data $\mathcal{I}$ and accomplishes two goals: authentication and identification.
For the former goal, we employ EEG signals to justify the impostor for its high fake-resistance. EEG signals are invisible and unique, making them difficult to be duplicated and hacked. 
For the latter goal, we adopt a deep learning model to extract the distinctive features and feed them into a non-parametric neighbor-based classifier for ID identification. In summary, the Deepkey authentication algorithm contains several key stages, namely {\em Invalid ID Filter}, {\em Gait-based Identification}, {\em EEG-based Identification} and {\em Decision Making}.
The overall authentication contains the following several stages (Figure~\ref{fig:attention}):
\begin{enumerate}
\item Based on the EEG data, the Invalid ID Filter decides the subject is an impostor or a genuine. If the subject is an impostor, the request will be denied.
\item If the individual is determined as genuine, the EEG/Gait Identification Model will identify the individual's authorized EEG/Gait ID. This model is pre-trained off-line with the attention-based Long Short-Term Member (LSTM) model (Section~\ref{sec:eeg_id_identification_model}). The output is the ID number associated with the person's detailed personal information. 
\item The final stage is to check the consistency of the EEG ID and the Gait ID. If they are identical,
the system will grant an approval, otherwise, deny the subject and take corresponding security measures.
\end{enumerate}

\begin{table}%
\caption{Notation}
\label{tab:notation}
\begin{minipage}{\columnwidth}
\begin{center}
\begin{tabular}{ll}
  \toprule
  \bf Parameters& \bf Explanation\\
  \hline
  $L^{\circ}$     & the number of authorized subjects (genuine)\\
  $E$       & the set of EEG signals \\  
  $G$       & the set of gait signals \\ 
  $n_s$     & the number of features in input data sample\\
  $G_i$     & the $i$-th gait data\\
  $n_g$     & the number of features in per gait sample\\
  $E_i$     & the $i$-th EEG data\\
  $n_e$     & the number of features in EEG data sample\\
  $X^i$     & the data in the $i$-th layer of attention RNN\\
  $N^i$     &the number of dimensions in $X^i$\\
  $K$       & the number of participants (genuine and impostor) \\
  $c^{i'}_{j} $ & the hidden state in the $j$-th LSTM cell \\
  $\mathcal{T}(\cdot)$ & the linear calculation of dense neural layers \\
  $\mathcal{L}(\cdot)$ & The LSTM calculation process \\
  $\odot$ &  the element\-wise multiplication \\ 
  $C_{att}$ &  attention based code \\
  \bottomrule
\end{tabular}
\end{center}
\end{minipage}
\end{table}%

\begin{figure}[t]
  \includegraphics[width=0.8\linewidth]{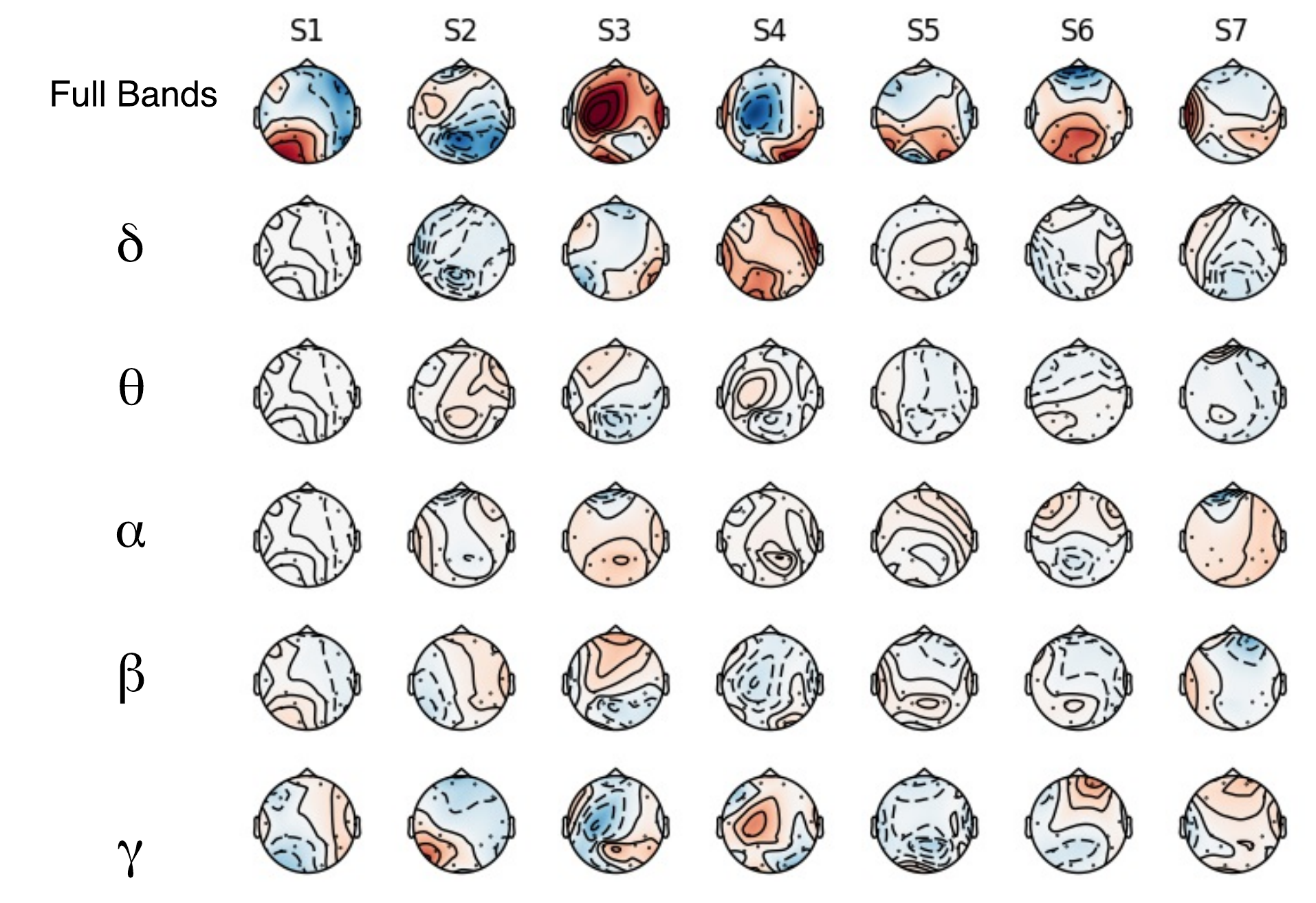}
  \caption{EEG topography of different subjects under different frequency bands. The inter-subject EEG signal cosine-similarity is calculated under each band and the results are reported as $0.1313$ (full bands), $0.0722$ (Delta band), $0.1672$ (Theta band), $0.2819$ (Alpha band), $0.0888$ (Beta band), and $0.082$ (Gamma band). This illustrates that the delta band with the lowest similarity contains the most distinguishable features for person identification.}
  \label{fig:topo}
\end{figure}

\subsection{Invalid ID Filter Model} 
\label{sec:invalid_detection_model}
Since the high fake-resistance of EEG data and rich distinguishable features in EEG signal, which lead to competitive invalid filter performance, in the Invalid ID Filter Model, only the EEG data $E_i$ is utilized to detect the genuine\footnote{Through our preliminary experiments, the gait signal is not effectiveness enough for invalid ID filtering.}. 

The subjects in an authentication system are categorized into two classes: {\em authorized} and {\em unauthorized}. Since the unauthorized data are not available in training stage, therefore, an unsupervised learning algorithm is required to identify the invalid ID. In this work, we apply one-class SVM to sort out the unauthorized subjects. 
Given a set of authorized subjects $\mathcal{S}=\left \{S_{i},i=1,2,\cdots,L^{\circ}\right \}$, 
$S_{i}\in R^{n_s}$, where $L^{\circ}$ denotes the number of authorized subjects and $n_s$ denotes the number of dimensions 
of the input data.
The input data consist of  
EEG data $\mathcal{E}=\left \{E_{i},i=1,2,\cdots,L^{\circ}\right \}$, $E_{i}\in R^{n_e}$ and gait data $\mathcal{G}=\left \{  G_{i},i=1,2,\cdots,L^{\circ}\right \}, G_{i}\in R^{n_g}$. 
$n_g$ and $n_e$ denote the number of dimensions 
of the 
gait data and EEG data, respectively, and 
$n_s=n_g+n_e$. The notation can be found in Table~\ref{tab:notation}.

For each authentication, the collected EEG data $E_i$ includes a number of samples. Each sample is a vector with shape $[1,14]$ where $14$ denotes the number of electric-nodes in Emotiv headset. To trade-off the authentication efficiency (less collecting and waiting time) and 
computational performance, 
based on the experimental experience, we fed 200 samples ([200, 14]) into the Invalid ID Filter. 200 EEG samples are collected in 1.56 seconds which is acceptable.
The final filter result is the mean of the results on all the samples. 

\begin{table*}[!tb]
\centering
\caption{Characteristics of EEG frequency bands. Awareness Degree denotes the degree of being aware of an external world.}
\label{tab:bands}
\resizebox{\textwidth}{!}{
\begin{tabular}{llllll}
\hline
\textbf{Name} & \textbf{Frequency ($Hz$)} & \textbf{Amplitude} & \textbf{Brain State} & \textbf{Awareness Degree} & \textbf{Produced Location} \\ \hline
\textbf{Delta} & 0.5-3.5 & Higher & Deep sleep pattern & Lower & Frontally and posteriorly \\
\textbf{Theta} & 4-8 & High & Light sleep pattern & Low &  Entorhinal cortex, hippocampus  \\
\textbf{Alpha} & 8-12 & Medium & Closing the eyes, relax state & Medium & Posterior regions of head \\
\textbf{Beta} & 12-30 & Low & \begin{tabular}[c]{@{}l@{}}Active thinking, focus, \\high alert, anxious \end{tabular} & High & Most evident frontally \\
\textbf{Gamma} & 30-100 & Lower & \begin{tabular}[c]{@{}l@{}}During cross-modal \\sensory processing \end{tabular} & Higher & Somatosensory cortex\\ \hline
\end{tabular}
}
\end{table*}

\subsection{EEG Identification Model} 
\label{sec:eeg_id_identification_model}
Compared to 
gait data, 
EEG data contain more noise which is more challenging to handle. Given the complexity of EEG signals, the data pre-processing is necessary. In practical EEG data analysis, the assembled EEG signals can be divided into several different frequency patterns (delta, theta, alpha, beta, and gamma) based on the strong intra-band correlation with a distinct behavioral state. 
The EEG frequency patterns and the corresponding characters are listed in Table~\ref{tab:bands} \cite{zhang2018mindid}. Figure~\ref{fig:topo} reports the topography of EEG signals of different subjects under different frequency bands and demonstrates that the Delta wave, compared to other bands, enriches distinctive features. In detail, we calculate the inter-subject EEG signal cosine-similarity which measure the average similarity among different subjects under all the EEG bands. The results are reported as 0.1313 (full bands), 0.0722(Delta band), 0.1672 (Theta band), 0.2819 (Alpha band), 0.0888 (Beta band), and 0.082 (Gamma band). This
illustrates that the delta band with the lowest similarity contains the most distinguishable features for person
identification.  Our previous work \cite{zhang2018mindid} has demonstrated that Delta pattern, compare to other EEG patterns contains the most distinctive information and is the most stable pattern in different environments by qualitative analysis and empirical experiment results. Thus, in this paper, we adopt a bandpass (0.5Hz-3.5Hz) butter-worth filter to extract Delta wave signal for further authentication. For simplicity, we denote the filtered EEG data as $\mathcal{E}$.

Since different EEG channels record different aspects of the brain signals, some of which are more representative of the individual, an approach that assumes all dimensions to be equal may not be suitable. Thus, we attempt to develop a novel model which can pay more attention to the most informative signals. In particular, the proposed approach is supposed to automatically learn the importance of the different parts of the EEG signal and focus on the valuable part. The effectiveness of attention-based RNN has been demonstrated in various domains including natural language processing \cite{wang2016attention} and speech recognition \cite{chan2017speech}. 
Inspired by the wide success of this approach, we introduce the attention mechanism to the Encoder-Decoder RNN model to assign varying weights to different dimensions of the EEG data.
After EEG filtering, the composed Delta pattern $\mathcal{E}$ is fed into an attention-based Encoder-Decoder RNN structure \cite{wang2016attention} aiming to learn more representative features for user identification. The general Encoder-Decoder RNN framework regards all the feature dimensions of input sequence has the same weights, no matter how important the dimension is for the output sequence. In this paper, the different feature dimensions of the EEG data 
 correspond 
 to the different nodes of the EEG equipment. For example, the first dimension (first channel) collects the EEG data from the $AF3$\footnote{Both $AF3$ and $O1$ are EEG measurement positions in the International 10-20 Systems.} node 
 located at the frontal lobe of the scalp while the 7-th dimension is gathered from $O1$ node at the occipital lobe. 
 To assign 
 various 
 weights to different dimensions of $\mathcal{E}$, we introduce the attention mechanism to the Encoder-Decoder RNN model. The proposed attention-based Encoder-Decoder RNN 
 consists of three components (Figure~\ref{fig:attention}): the encoder, the attention module, and the decoder. The encoder is designed to compress the input Delta $\delta$ wave into a single intermediate code $C$; the attention module helps the encoder 
 calculate a better intermediate code $C_{att}$ by generating a sequence of the weights $W_{att}$ of different dimensions; the decoder accepts the attention-based code $C_{att}$ and decodes it to the output layer $Y'$. 

Suppose the data in $i$-th layer can be denoted by $X^i=(X^i_j;i\in[1,2,\cdots,I], j\in[1,2,\cdots, N^i])$
where $j$ denotes the $j$-th dimension of $X^i$. $I$ represents the number of neural network layers while $N^i$ denotes the number of dimensions in $X^i$. Take the first layer as an example, we have $X^1=\mathcal{E}$ which indicates that the input sequence is the Delta pattern. Let the output sequence be $Y=(Y_k; k\in[1,2,\cdots,K])$ where K denotes the number of users.
In this paper, the user ID is represented by the one-hot label with length $K$. 
For simplicity, we define the operation $\mathcal{T}(\cdot)$ as:$\mathcal{T}(X^i)=X^iW+b$. 
Further more, we have
$$\mathcal{T}(X^{i-1}_j,X^i_{j-1})=X^{i-1}_j*W'+X^i_{j-1}*W''+b'$$
where $W$, $b$, $W'$, $W''$, $b'$ denote the corresponding weights and biases parameters. 

The encoder component contains several non-recurrent fully-connected layers and one recurrent Long Short-Term Memory (LSTM) layer. The non-recurrent layers are employed to construct and fit into a non-linear function to purify the input Delta pattern, the necessity is demonstrated by the preliminary experiments\footnote{Some optimal designs like the neural network layers are validated by the preliminary experiments but the validation procedure will not be reported in this paper for space limitation.}. 
The data flow in these non-recurrent layers can be calculated by 
$$X^{i+1}=tanh(\mathcal{T}(X^i))$$
where $tanh$ is the activation function. We engage the $tanh$ as an activation function instead of $sigmoid$ for the stronger gradient \cite{lecun1998efficient}. The LSTM layer is adopted to compress the output of non-recurrent layers to a length-fixed sequence which is regarded as the intermediate code $C$. Suppose LSTM is the $i'$-th layer, the code equals to the output of LSTM, which is $C=X^{i'}_j$. The $X^{i'}_j$ can be measured by
\begin{equation}
\label{equ:1}
X^{i'}_j=\mathcal{L}(c^{i'}_{j-1},X^{i-1}_j,X^{i'}_{j-1})
\end{equation}
 where $c^{i'}_{j-1}$ denotes the hidden state of the $(j-1)$-th LSTM cell. The operation $\mathcal{L}(\cdot)$ denotes the calculation process of the LSTM structure, which can be inferred from the following equations.
 \[X^{i'}_{j}=f_o\odot tanh(c^{i'}_{j})\]
  \[c^{i'}_{j}=f_f\odot c^{i'}_{j-1}+f_i\odot f_m\]
 \[f_o=sigmoid(\mathcal{T}(X^{i'-1}_{j},X^{i'}_{j-1}))\]
\[f_f=sigmoid(\mathcal{T}(X^{i'-1}_{j},X^{i'}_{j-1}))\]
 \[f_i=sigmoid(\mathcal{T}(X^{i'-1}_{j},X^{i'}_{j-1}))\]
\[f_m=tanh(\mathcal{T}(X^{i'-1}_{j},X^{i'}_{j-1}))\]
where $f_o,f_f, f_i$ and $f_m$ represent the output gate, forget gate, input gate and input modulation gate
,
respectively, and $\odot$ denotes the element-wise multiplication. 

The attention module accepts the final hidden states as the unnormalized attention weights $W'_{att}$ which can be measured by the mapping operation $\mathcal{L}'(\cdot)$ (similar with Equation~\ref{equ:1})
$$W'_{att}=\mathcal{L}'(c^{i'}_{j-1},X^{i-1}_j,X^{i'}_{j-1})$$
and calculate the normalized attention weights $W_{att}$ 
$$W_{att}=softmax(W'_{att})$$
The softmax function is employed to normalize the attention weights into the range of $[0,1]$. Therefore, the weights can be explained as the probability that how the code $C$ is relevant to the output results. 
Under the attention mechanism, the code $C$ is weighted to $C_{att}$ 
$$C_{att}=C\odot W_{att}$$
Note, $C$ and $W_{att}$ are trained instantaneously.
The decoder receives the attention-based code $C_{att}$ and decodes it to the output $Y'$.
Since $Y'$ is predicted at the output layer of the attention based RNN model ($Y'=X^I$), we have
$$Y'=\mathcal{T}(C_{att})$$
At last, we employ the cross-entropy cost function and $\ell_2$-norm (with parameter $\lambda$) is selected to prevent overfitting. The cost is optimized by the AdamOptimizer algorithm \cite{kingma2014adam}. 
The iterations threshold of attention-based RNN is set as $n^E_{iter}$. 
The weighted code $C_{att}$ has a 
direct linear relationship with the output layer and the predicted results. If the model is well trained with low cost, we could regard the weighted code as a high-quality representation of the user ID. We set the learned deep feature $X_D$ to $C_{att}$, $X_D=C_{att}$, and feed it into a lightweight nearest neighbor classifier. The EEG ID, which is denoted by $E_{ID}$, can be directly predicted by the classifier.

\begin{algorithm}[!t]
\caption{Deepkey System}
\label{alg:Deepkey}
\begin{algorithmic}[1]
\renewcommand{\algorithmicrequire}{\textbf{Input:}}
\renewcommand{\algorithmicensure}{\textbf{Output:}}
 \REQUIRE EEG data $\mathcal{E}$ and Gait data $\mathcal{G}$
 \ENSURE  Authentication Decision: Approve/Deny 
 \STATE \#Invalid ID Filter:
 \FOR{$\mathcal{E}, \mathcal{G}$} 
    \STATE Genuine/Impostor $\gets \mathcal{E}$ 
    \IF{Impostor}
        \RETURN Deny
    \ELSIF{Genuine}
        \STATE \#EEG Identification Model:
        \WHILE{$iteration<n^E_{iter}$}
            \STATE $X^{i+1}=tanh(\mathcal{T}(X^i))$ \COMMENT{$X^1 = \mathcal{E}$}
            \STATE $C=X^{i'}_j=\mathcal{L}(c^{i'}_{j-1},X^{i-1}_j,X^{i'}_{j-1})$
            \STATE $W_{att}=softmax(\mathcal{L}'(c^{i'}_{j-1},X^{i-1}_j,X^{i'}_{j-1}))$
            \STATE $C_{att}=C\odot W_{att}$
            \STATE $E_{ID}\gets C_{att}$
        \ENDWHILE
        \STATE \#Gait Identification Model:
        \WHILE{$iteration<n^G_{iter}$}
            \STATE $G_{ID}\gets \mathcal{G}$
        \ENDWHILE
        \IF{$E_{ID}=G_{ID}$}
            \RETURN Approve
        \ELSE
            \RETURN Deny
        \ENDIF
    \ENDIF
\ENDFOR
\end{algorithmic}
\end{algorithm}

The Gait Identification Model 
works similar as the EEG Identification Model except the frequency band filtering. The iterations threshold of attention-based RNN is set as $n^G_{iter}$. 
The Gait Identification Model receives subjects' gait data $\mathcal{G}$ from the input data $\mathcal{I}$ and 
maps to the user's Gait ID $G_{ID}$. All the model structures, hyper-parameters, optimization, and other settings in the EEG and Gait Identification Models remain the same to keep the lower model complexity of the Deepkey system.

\begin{figure}[!t]
  \includegraphics[width=\linewidth]{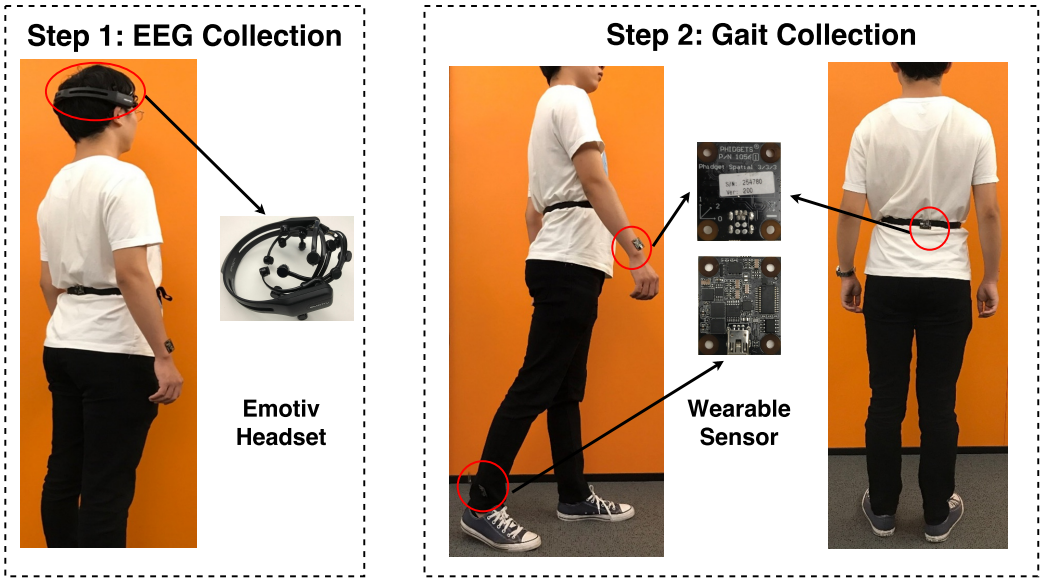}
  \caption{Data collection. Two collection steps are cascade to eliminate the impact on EEG data of walking. The first step collects the solo EEG signals while the second step collects the gait signals.}
  \label{fig:collection}
\end{figure}

\section{Experiments and results} 
\label{sec:experiment_and_results}
In this section, we first outline the experimental setting including dataset, hyper-parameters settings, and evaluation metrics. Then we systematically investigate:
1) the comparison with the state-of-the-art authentication systems in both system-level and component-level; 
2) the impact of key paramters like single/multiple sessions\footnote{Single session refers to the dataset collected in one session (the period from one subject putting the EEG headset on until all the experiments are finished then putting off). Multi-session represents the EEG data is collected from different sessions, which considered the effect on EEG data quality caused by the headset position errors.}, EEG band, and datasize;
3) the authentication latency?

\subsection{Experimental Settings} 
\label{sub:setting}

\begin{table}[!bt]
\centering
\caption{Datasets description. \#-D denotes the number of dimensions.}
\label{tab:dataset}
\resizebox{0.6\textwidth}{!}{
\begin{tabular}{llllll}
\rowcolor[HTML]{FFFFFF} 
\hline
\textbf{Dataset} & \textbf{Biometric} & \textbf{\#-D} & \textbf{Session} & \textbf{Frequency}  & \textbf{Samples} \\ \hline
\textbf{EID-S} & EEG & 14 & Single & 128Hz  & 49,000 \\
\textbf{EID-M} & EEG & 14 & Multiple & 128Hz  & 147,000 \\
\textbf{GID-S} & Gait & 27 & Single & 80Hz  & 140,000 \\
\textbf{GID-M} & Gait & 27 & Multiple & 80Hz  & 420,000 \\ \hline
\end{tabular}
}
\vspace{-3mm}
\end{table}

\begin{table*}[!bt]
\centering
\caption{System-level comparison between Deepkey and other biometrics authentication systems. The performance of our methods are evaluated on multi-session datasets (EID-M, GID-M).}
\label{tab:system_comparison}
\resizebox{\textwidth}{!}{\begin{tabular}{lllllllll}
\hline \hline
 & \textbf{Reference} & \textbf{Biometric} & \textbf{Method} & \#-Subject & Dataset & \textbf{Accuracy}  & \textbf{FAR}  & \textbf{FRR}  \\ \hline
\multirow{11}{*}{\begin{tabular}[c]{@{}l@{}}\textbf{Uni-}\\ \textbf{-modal}\end{tabular}} & \cite{cola2016gait} & Gait & semi-supervised anomaly detection+NN & 15 & Local & 97.4 &  &   \\
 & \cite{muramatsu2015gait} & Gait & AVTM-PdVs &100  & Public & 77.72 &  &   \\
 & \cite{al2017unobtrusive} & Gait & MLP & 60 &Local & 99.01 &  &   \\
 & \cite{sun2014gait} & Gait & Artificial features + voting classifier & 10 &Local & 98.75 &  &   \\
 & \cite{konno2015gait} & Gait & Two SVMs & 50 &Local &  & 1.0 & 1.0  \\
 & \cite{thomas2017eeg} & EEG & PSD + cross-correlation values &109  &Public &  & 1.96 &1.96   \\
  & \cite{chuang2013think} & EEG & Customized Threshold &15  &Local &  &  0& 2.2 \\
  & \cite{gui2014exploring} & EEG &Low-pass filter+wavelets+ ANN  &32  &Local & 90.03 &  &  \\
 & \cite{bashar2016human} & EEG & Bandpass FIR filter +ECOC + SVM & 9 & Local & 94.44 &  &   \\
 & \cite{thomas2016utilizing} & EEG & \begin{tabular}[c]{@{}l@{}}IAF + delta band EEG  \\ + Cross-correlation values\end{tabular} &109  &Public & 90.21 &  &   \\
 & \cite{jayarathne2016brainid} & EEG & CSP +LDA &  12 & Local & 96.97 & &   \\
 & \cite{zhang2018mindid} & EEG & Attention-based RNN + XGB & 8 &Local & 98.82 &  &   \\
 & \cite{keshishzadeh2016improved} & EEG & AR + SVM & 104 & Public& 97.43 &  &   \\ \hline \hline
\multirow{18}{*}{\begin{tabular}[c]{@{}l@{}}\textbf{Multi-}\\ \textbf{-modal}\end{tabular}} & \multirow{3}{*}{\cite{long2012multimodal}} & Fingerprint & \multirow{3}{*}{ZM + RBF Neural Network} &  & & 92.89 & 7.108 & 7.151  \\
 &  & Face &  &  & &   & 11.52 & 13.47 \\
 &  & Fusion &  & 40 &Public &  & 4.95 & 1.12 \\ \cline{2-9} 
 & \multirow{3}{*}{\cite{yano2012multimodal}} & Iris & \multirow{3}{*}{\begin{tabular}[c]{@{}l@{}}Gabor 2D wavelets + Gabor 2D wavelets\\  + Hamming distance \end{tabular}} &  & &  & 13.88 &  13.88 \\
 &  & Pupil &  &  & &  &5.47  & 5.47 \\
 &  & Fusion &  & 59 &Public &  &2.44  & 2.44  \\ \cline{2-9} 
 & \multirow{3}{*}{\cite{manjunathswamy2015multimodal}} & ECG & wavelet decomposition &  & &  & 2.37 & 9.52  \\
 &  & Fingerprint & Histogram manipulation Image Enhancement &  & &  & 7.77 & 5.55  \\
 &  & Fusion & Score Fusion &50  &Public &  & 2.5 & 0   \\ \cline{2-9} 
 & \multirow{3}{*}{\cite{derawi2014fusion}} & ECG & \multirow{2}{*}{linear time interpolation + cross correlation} &   & & & 4.2 & 4.2   \\
 &  & Gait &  &  & &  & 7.5 & 7.5 \\
 &  & Fusion & Score Fusion & 30 &Local &  & 1.26 &1.26   \\ \cline{2-9} 
 & \multirow{3}{*}{Ours} & EEG & Delta wave + Attention-based RNN +KNN &  & & 99.96 &  &   \\
 &  & Gait & Attention-based RNN +KNN &  & & 99.61 &  & \\
 &  & Fusion &  &7  &Local & \textbf{99.57} & \textbf{0} & \textbf{1.0}  \\ \hline \hline
\end{tabular}}
\end{table*}

\begin{figure*}[t]
\hspace{5mm}
    \centering
    \begin{subfigure}[t]{0.25\textwidth}
    \centering
    \includegraphics[width=\textwidth]{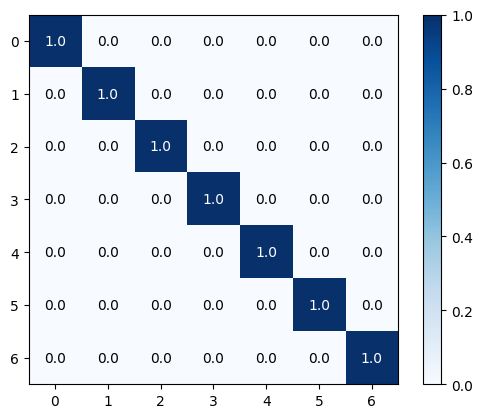}
    \caption{EID-S CM}
    \end{subfigure}%
    \begin{subfigure}[t]{0.25\textwidth}
    \centering
    \includegraphics[width=\textwidth]{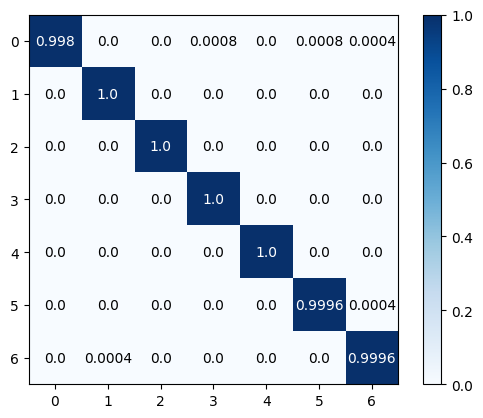}
    \caption{EID-M CM}
    \end{subfigure}%
    \begin{subfigure}[t]{0.25\textwidth}
    \centering
    \includegraphics[width=\textwidth]{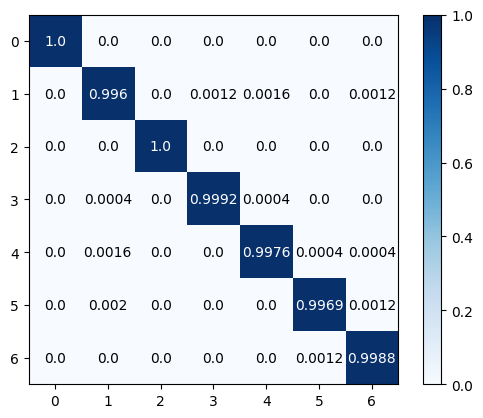} 
    \caption{GID-S CM}
    \end{subfigure}%
    \begin{subfigure}[t]{0.25\textwidth}
    \centering
    \includegraphics[width=\textwidth]{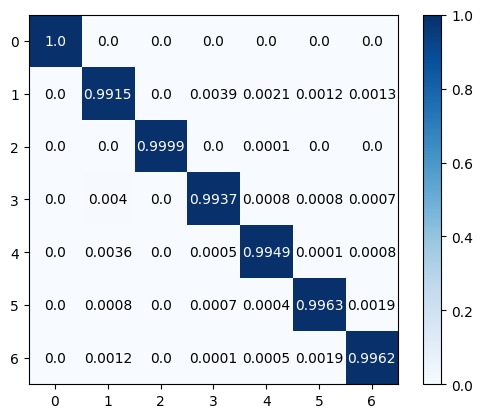} 
    \caption{GID-M CM}
    \end{subfigure}%

    \centering
    \begin{subfigure}[t]{0.25\textwidth}
    \centering
    \includegraphics[width=\textwidth]{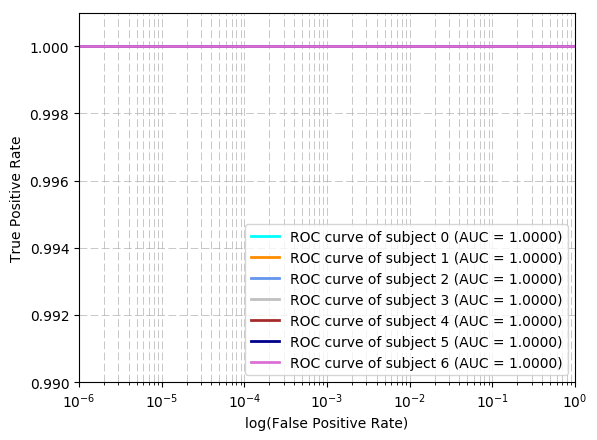}
    \caption{EID-S ROC}
    \end{subfigure}%
    \begin{subfigure}[t]{0.25\textwidth}
    \centering
    \includegraphics[width=\textwidth]{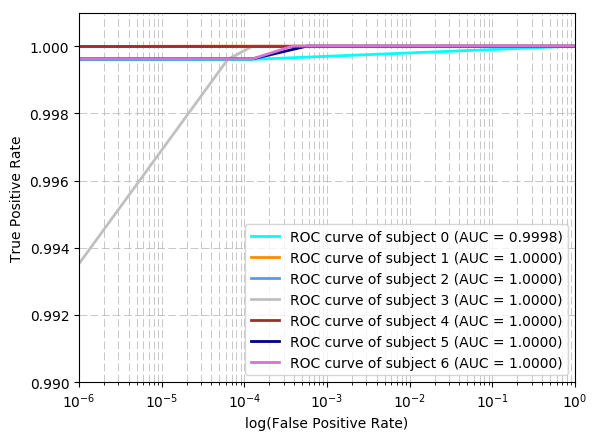}
    \caption{EID-M ROC}
    \end{subfigure}%
    \begin{subfigure}[t]{0.25\textwidth}
    \centering
    \includegraphics[width=\textwidth]{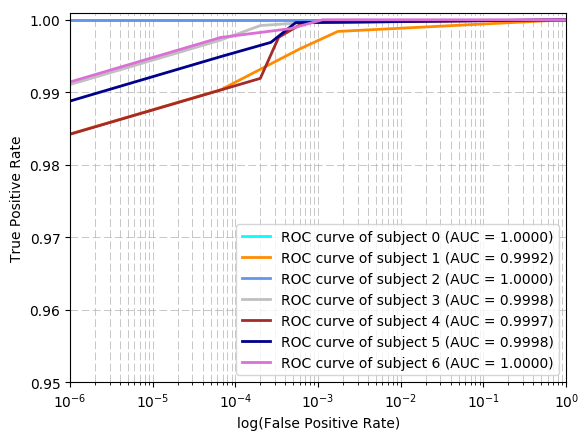}
    \caption{GID-S ROC}
    \end{subfigure}%
    \begin{subfigure}[t]{0.25\textwidth}
    \centering
    \includegraphics[width=\textwidth]{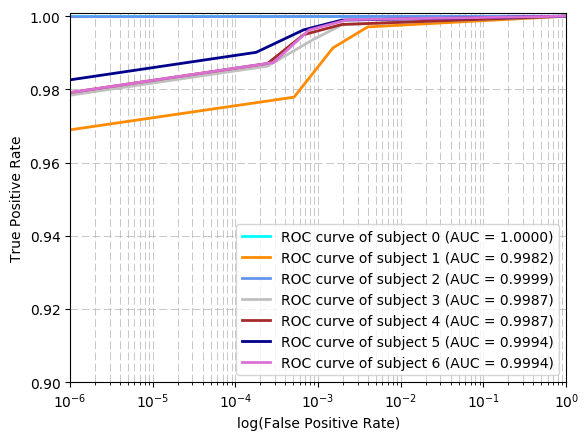}
    \caption{GID-M ROC}
    \end{subfigure}%
    \caption{Confusion matrix and ROC curves of the datasets. CM denotes confusion matrix. The AUC are provided on the figures. }
    \label{fig:cm_roc}
\end{figure*}

\begin{table*}[tb]
\centering
\caption{Classification report of the datasets including precision, recall and F-1 score. The proposed approach gains impressive results (higher than 99\%) on all the metrics over all the 7 subjects.
 }
\label{tab:classification_report}
\resizebox{\textwidth}{!}{
\begin{tabular}{l|lll|lll|lll|lll}
\hline
\textbf{Datasets} & \multicolumn{3}{c|}{\textbf{EID-S}} & \multicolumn{3}{c|}{\textbf{EID-M}} & \multicolumn{3}{c|}{\textbf{GID-S}} & \multicolumn{3}{c}{\textbf{GID-M}} \\ \hline
\textbf{Metrics} & \textbf{Precision} & \textbf{Recall} & \textbf{F1-score} & \textbf{Precision} & \textbf{Recall} & \textbf{F1-score} & \textbf{Precision} & \textbf{Recall} & \textbf{F1-score} & \textbf{Precision} & \textbf{Recall} & \textbf{F1-score} \\ \hline
\textbf{0} & 1.0 & 1.0 & 1.0 & 0.998 & 1.0 & 0.999 & 1.0 & 1.0 & 1.0 & 1.0 & 1.0 & 1.0 \\
\textbf{1} & 1.0 & 1.0 & 1.0 & 1.0 & 0.9996 & 0.9998 & 0.996 & 0.996 & 0.996 & 0.9915 & 0.9904 & 0.9909 \\
\textbf{2} & 1.0 & 1.0 & 1.0 & 1.0 & 1.0 & 1.0 & 1.0 & 1.0 & 1.0 & 0.9999 & 1.0 & 0.9999 \\
\textbf{3} & 1.0 & 1.0 & 1.0 & 1.0 & 0.9992 & 0.9996 & 0.9992 & 0.9988 & 0.999 & 0.9937 & 0.9948 & 0.9942 \\
\textbf{4} & 1.0 & 1.0 & 1.0 & 1.0 & 1.0 & 1.0 & 0.9976 & 0.998 & 0.9978 & 0.9949 & 0.996 & 0.9955 \\
\textbf{5} & 1.0 & 1.0 & 1.0 & 0.9996 & 0.9992 & 0.9994 & 0.9969 & 0.9984 & 0.9977 & 0.9963 & 0.996 & 0.9961 \\
\textbf{6} & 1.0 & 1.0 & 1.0 & 0.9996 & 0.9993 & 0.9994 & 0.9988 & 0.9972 & 0.998 & 0.9962 & 0.9953 & 0.9958 \\
\textbf{Average} & \textbf{1.0} & \textbf{1.0} & \textbf{1.0} & \textbf{0.9996} & \textbf{0.9996} & \textbf{0.9996} & \textbf{0.9983} & \textbf{0.9983} & \textbf{0.9983} & \textbf{0.9961} & \textbf{0.9961} & \textbf{0.9961} \\ \hline
\end{tabular}
}
\end{table*}

\subsubsection{Datasets} 
\label{sub:datasets}

We design real-world experiments to collect EEG data and gait data in cascade. The experiments (Figure~\ref{fig:collection}) are conducted by 7 healthy participants aged $26\pm2$ including 4 males and 3 females. 
In Step 1, each participant remains standing and 
relaxed with eye closed. 
The EEG data are collected by EPOC+ Emotiv headset\footnote{\url{https://www.emotiv.com/product/emotiv-epoc-14-channel-mobile-eeg/}} which integrates 14 electrodes (corresponding to 14 EEG channels) with 128Hz sampling rate. 
In Step 2, each participant walks in an aisle to generate the gait data. In the gait collection procedure, three IMUs are attached to the participants' left wrist, the middle of the back, and the left ankle, respectively. Each IMU (PhidgetSpatial 3/3/3\footnote{\url{https://www.phidgets.com/?&prodid=48}}) with 80Hz sampling rate gathering 9 dimensional motor features contains a 3-axis accelerometers, 3-axis gyroscopes, and 3-axis magnetometers. 

To investigate the impact of dataset sessions, both the EEG and gait data are collected in three sessions. In every cycle of single session, the subject puts on the equipment (headset/IMUs), gathers data, and then takes the equipment off. Therefore, in different sessions, the positions of the equipment may have slight deviation. 
Table~\ref{tab:dataset} reports the details of the datasets used in this paper. Each EEG or gait sample contains 10 continuous instances without overlapping. 
The single session datasets (EID-S and GID-S) are collected in single experiment session whilst the multi-session datasets (EID-M and GID-M) are gathered in three sessions. All the sessions are conducted in the same place but three different days (each session in one day). The EEG data are easily influenced if the emotional or physical states has changed, thus, we believe the collected data are diverse because of the varying environmental factors (e.g., noise and temperature) and subjective factors (e.g., participants' mental state and fatigue state). Similarly, the gait signals could be affected by lots of variables like different shoes (comfortable/uncomfortable). In this paper, as an exploratory work, we focus to develop a robust discriminative deep learning model which is strong enough to prevent the corruption of the aforementioned influencing factors. The investigation of the detail effect brought by each specific factor will be left as a future research direction. 

\subsubsection{Parameter Settings} 
\label{sub:parameter_settings}
The Invalid ID Filter attempts to recognize the unauthorized subject based on the unique EEG data. The filter chooses the RBF kernel with $nu = 0.15$. In the EEG Identification Model, the Delta band ($[0.5Hz,3.5Hz]$) is filtered by 3 order butter-worth bandpass filter. In the attention-based RNN, for both EEG and gait, each hidden layer contains 64 nodes, while the learning rate and $\lambda$ are both set to 0.001. The weighted code is produced after 1,000 iterations. 
87.5\% of the datasets are used for training while the remaining datasets are used for testing. The user ID is ranged from 0 to 6 and represented in the one-hot label.

\begin{table*}[t]
\centering
\caption{Component-level comparison. DL denotes Deep Learning. Att-RNN denotes attention-based RNN.
}
\label{tab:component_level_comparison}
\resizebox{\textwidth}{!}{
\begin{tabular}{l|l|llll|llll}
\hline \hline
\multirow{2}{*}{\textbf{Baseline}} & \multirow{2}{*}{\textbf{Methods}} & \multicolumn{4}{c|}{\textbf{EID-S}} & \multicolumn{4}{c}{\textbf{EID-M}} \\ \cline{3-10} 
 &  & \textbf{Accuracy} & \textbf{Precision} & \textbf{Recall} & \textbf{F1-score} & \textbf{Accuracy} & \textbf{Precision} & \textbf{Recall} & \textbf{F1-score} \\ \hline
\multirow{5}{*}{\textbf{Non-DL Baseline}} & \textbf{SVM} & 0.4588 & 0.5848 & 0.4588 & 0.4681 & 0.7796 & 0.7815 & 0.7796 & 0.7796 \\
 & \textbf{RF} & 0.9875 & 0.9879 & 0.9876 & 0.9876 & 0.8124 & 0.8139 & 0.8124 & 0.812 \\
 & \textbf{KNN} & 0.9897 & 0.9899 & 0.9898 & 0.9898 & 0.8211 & 0.8232 & 0.8211 & 0.8197 \\
 & \textbf{AB} & 0.2872 & 0.3522 & 0.2871 & 0.2337 & 0.3228 & 0.3224 & 0.3228 & 0.2815 \\
 & \textbf{LDA} & 0.1567 & 0.1347 & 0.1567 & 0.1386 & 0.3082 & 0.285 & 0.3082 & 0.2877 \\ \hline
\multirow{3}{*}{\textbf{DL Baseline}} & \textbf{LSTM} & 0.9596 & 0.9601 & 0.9596 & 0.9597 & 0.8482 & 0.8509 & 0.8483 & 0.8489 \\
 & \textbf{GRU} & 0.9633 & 0.9636 & 0.99631 & 0.9631 & 0.862 & 0.8638 & 0.8626 & 0.8629 \\
 & \textbf{CNN} & 0.8822 & 0.8912 & 0.8813 & 0.8912 & 0.7647 & 0.7731 & 0.7854 & 0.7625 \\ \hline
\multirow{3}{*}{\textbf{State-of-the-art}} & \textbf{\cite{jayarathne2016brainid}} & 0.5843 & 0.5726 & 0.5531 & 0.5627 & 0.5735 & 0.5721 & 0.5443 & 0.5579 \\
 & \textbf{\cite{keshishzadeh2016improved}} & 0.8254 & 0.8435 & 0.8617 & 0.8525 & 0.8029 & 0.7986 & 0.8125 & 0.8055 \\
 & \textbf{\cite{gui2014exploring}} & 0.8711 & 0.8217 & 0.7998 & 0.8106 & 0.8567 & 0.8533 & 0.8651 & 0.8592 \\ \hline
\textbf{} & \textbf{Att-RNN} & 0.9384 & 0.9405 & 0.9388 & 0.9391 & 0.9324 & 0.9343 & 0.9322 & 0.9326 \\
\textbf{} & \textbf{Ours} & \textbf{1.0} & \textbf{1.0} & \textbf{1.0} & \textbf{1.0} & \textbf{0.9996} & \textbf{0.9996} & \textbf{0.9996} & \textbf{0.9996} \\ \hline \hline
\multirow{2}{*}{\textbf{Baselines}} & \multirow{2}{*}{\textbf{Methods}} & \multicolumn{4}{c|}{\textbf{GID-S}} & \multicolumn{4}{c}{\textbf{GID-M}} \\ \cline{3-10} 
 &  & \textbf{Accuracy} & \textbf{Precision} & \textbf{Recall} & \textbf{F1-score} & \textbf{Accuracy} & \textbf{Precision} & \textbf{Recall} & \textbf{F1-score} \\ \hline
\multirow{5}{*}{\textbf{Non-DL Baseline}} & \textbf{SVM} & 0.9981 & 0.9981 & 0.9981 & 0.9981 & 0.993 & 0.993 & 0.993 & 0.993 \\
\textbf{} & \textbf{RF} & 0.9878 & 0.9878 & 0.9878 & 0.9878 & 0.9954 & 0.9954 & 0.9954 & 0.9954 \\
\textbf{} & \textbf{KNN} & 0.9979 & 0.9979 & 0.9979 & 0.9979 & 0.9953 & 0.9953 & 0.9953 & 0.9953 \\
\textbf{} & \textbf{AB} & 0.5408 & 0.5689 & 0.5409 & 0.4849 & 0.5401 & 0.5135 & 0.542 & 0.4985 \\
\textbf{} & \textbf{LDA} & 0.688 & 0.6893 & 0.688 & 0.6855 & 0.6933 & 0.693 & 0.6933 & 0.6915 \\ \hline
\multirow{3}{*}{\textbf{DL Baseline}} & \textbf{LSTM} & 0.9951 & 0.9951 & 0.9951 & 0.9951 & 0.9935 & 0.9936 & 0.9936 & 0.9936 \\
\textbf{} & \textbf{GRU} & 0.9949 & 0.9949 & 0.9949 & 0.9949 & 0.9938 & 0.9938 & 0.9938 & 0.9938 \\
\textbf{} & \textbf{CNN} & 0.9932 & 0.9932 & 0.9932 & 0.9932 & 0.9845 & 0.9845 & 0.9845 & 0.9845 \\ \hline
\textbf{State-of-the-art} & \textbf{\cite{cola2016gait}} & 0.9721 & 0.9789 & 0.9745 & 0.9767 & 0.9653 & 0.9627 & 0.9669 & 0.9648 \\
\textbf{} & \textbf{\cite{al2017unobtrusive}} & 0.9931 & 0.9934 & 0.9957 & 0.9945 & 0.9901 & 0.9931 & 0.9942 & 0.9936 \\
\textbf{} & \textbf{\cite{sun2014gait}} & 0.9917 & 0.9899 & 0.9917 & 0.9908 & 0.9875 & 0.9826 & 0.9844 & 0.9835 \\ \hline
\textbf{} & \textbf{Att-RNN} & 0.99 & 0.99 & 0.99 & 0.99 & 0.9894 & 0.9895 & 0.9895 & 0.9895 \\
\textbf{} & \textbf{Ours} & \textbf{0.9983} & \textbf{0.9983} & \textbf{0.9983} & \textbf{0.9983} & \textbf{0.9961} & \textbf{0.9961} & \textbf{0.9961} & \textbf{0.9961} \\ \hline \hline
\end{tabular}
}
\end{table*}

\subsubsection{Metrics} 
\label{sub:evaluation}
The adopted evaluation metrics are accuracy, ROC, AUC, along with {\em False Acceptance Rate} (FAR) and {\em False Rejection Rate } (FRR).
Deepkey is very sensitive to the invalid subject and can acquire very high accuracy in Invalid ID Filter Model for the reason that even tiny misjudgment may lead to catastrophic consequences. Therefore, FAR is more important than other metrics. Therefore, in Deepkey, FAR has higher priority compared to other metrics such as FRR. 

\subsection{Overall Comparison} 
\label{sub:overall_comparison}

\subsubsection{System-level Comparison} 
\label{sub:system_level_comparison}
To evaluate the performance of Deepkey, by comparing it with a set of the state-of-the-art authentication systems. Deepkey is empowered to solve both the authentication and identification problems. As shown in Table~\ref{tab:system_comparison}, Deepkey achieves a FAR of 0 and a FRR of 1\%, outperforming other uni-modal and multimodal authentication systems. Specifically, our approach, compared to the listed uni-modal systems, achieves the highest EEG identification accuracy (99.96\%) and Gait identification accuracy (99.61\%). 

\subsubsection{Component-level Comparison} 
\label{sub:component_level_comparison}
To have a closer observation, we provide the detailed performance study of each component. 
In the Invalid ID Filter, to enhance the accuracy and robustness of the classifier, EEG samples are separated into different segments, with each segment (without overlapping) having $200$ continuous samples. 
Six in seven subjects are labeled as genuine while the other subject is labeled as 
an impostor.  In the training stage, all the EEG segments are fed into the one-class SVM with rbf kernel for pattern learning. In the test stage, 1,000 genuine segments and 1,000 impostor segments are randomly selected to asses the performance. We use the leave-one-out-cross-validation training strategy and achieve \bf{ a FAR of 0 and a FRR of 0.006}. 

In the EEG/Gait Identification Model, the proposed approach achieves an accuracy of 99.96\% and 99.61\% over the multi-session datasets, respectively. The detailed confusion matrix, ROC curves with AUC scores, and the classification reports (precision, recall, and F1-score) over all the datasets are presented in Figure~\ref{fig:cm_roc} and Table~\ref{tab:classification_report}. The above evaluation metrics demonstrate that the proposed approach achieves a performance of over 99\% on all the metrics over each subject and each dataset.

Furthermore, the overall comparison between our model and other state-of-the-art baselines
 are listed in Table~\ref{tab:component_level_comparison}. RF denotes Random Forest, AdaB denotes Adaptive Boosting, LDA denotes Linear Discriminant Analysis. In addition, the key parameters of the baselines are listed here: Linear SVM ($C=1$), RF ($n=200$), KNN ($k=3$). The settings of LSTM are the same as the attention-based RNN classifier, along with the GRU (Gated Recurrent Unit).
The CNN contains 2 stacked convolutional layers (both with stride $[1,1]$, patch $[2,2]$, zero-padding, and the depth are 4 and 8, separately.) and followed by one pooling layer (stride $[1,2]$, zero-padding) and one fully connected layer (164 nodes). Relu activation function is employed in the CNN.
The methods used for comparison (3 for EEG-based authentication and 3 for gait-based authentication) are introduced as follows:
\begin{itemize}
\item Jayarathne et al. \cite{jayarathne2016brainid} feed EEG data to a bandpass filter ($8Hz-30Hz$), extract CSP (Common Spatial Pattern) and recognize the user ID by LDA.
\item Keshishzadeh et al. \cite{keshishzadeh2016improved} extract autoregressive coefficients as the features and identify the subject by SVM.
\item Gui et al. \cite{gui2014exploring} employ low-pass filter (60Hz) and wavelet packet decomposition to generate features and distinguish unauthorized person through deep neural network. 
\item Cola et al. \cite{cola2016gait} 
hire neural networks to analyze user gait pattern by artificial features such as kurtosis, peak-to-peak amplitude and skewness.
\item Al-Naffakh et al. \cite{al2017unobtrusive} propose to utilize time-domain statistical features and Multilayer Perception (MLP) for person identify.
\item Sun et al. \cite{sun2014gait} adopt a weighted voting classifier to process the extracted features like gait frequency, symmetry coefficient, and dynamic range.
\end{itemize}

\begin{table*}[t]
\centering
\caption{EEG bands comparison. The full band denotes the raw EEG data with full frequency bands.}
\label{tab:bands_comparison}
\resizebox{\textwidth}{!}{
\begin{tabular}{lllllllllll}
\hline \hline
\multirow{2}{*}{\textbf{Dataset}} & \multirow{2}{*}{\textbf{Baseline}} & \multirow{2}{*}{\textbf{Methods}} & \multicolumn{6}{c}{\textbf{EEG Bands}} & \multirow{2}{*}{\textbf{Best Level}} & \multirow{2}{*}{\textbf{Best Band}} \\ \cline{4-9}
 &  &  & \textbf{Delta} & \textbf{Theta} & \textbf{Alpha} & \textbf{Beta} & \textbf{Gamma} & \textbf{Full} &  &  \\ \hline
\multirow{13}{*}{\textbf{EID-S}} & \multirow{5}{*}{\textbf{\begin{tabular}[c]{@{}l@{}}Non-DL\\  Baseline\end{tabular}}} & \textbf{SVM} & 0.4588 & 0.4682 & 0.7484 & 0.5955 & 0.5239 & 0.8788 & 0.8788 & Full \\
 &  & \textbf{RF} & 0.9875 & 0.8006 & 0.7729 & 0.6376 & 0.5469 & 0.8931 & 0.9875 & Delta \\
 &  & \textbf{KNN} & 0.9897 & 0.8084 & 0.7465 & 0.5553 & 0.4606 & 0.8792 & 0.9897 & Delta \\
 &  & \textbf{AB} & 0.2872 & 0.2879 & 0.2318 & 0.3069 & 0.2922 & 0.3289 & 0.3289 & Full \\
 &  & \textbf{LDA} & 0.1567 & 0.1802 & 0.1957 & 0.1502 & 0.1306 & 0.4547 & 0.4547 & Full \\ \cline{2-11} 
 & \multirow{3}{*}{\textbf{\begin{tabular}[c]{@{}l@{}}DL\\ Baseline\end{tabular}}} & \textbf{LSTM} & 0.9596 & 0.8126 & 0.8277 & 0.6906 & 0.6027 & 0.9273 & 0.9596 & Delta \\
 &  & \textbf{GRU} & 0.9633 & 0.7996 & 0.8082 & 0.6902 & \textbf{0.6985} & 0.9251 & 0.9633 & Delta \\
 &  & \textbf{CNN} & 0.8822 & 0.7416 & 0.8079 & \textbf{0.6918} & 0.6059 & 0.8985 & 0.8985 & Full \\ \cline{2-11} 
 &  & \textbf{\cite{jayarathne2016brainid}} & 0.5843 & 0.4487 & 0.2918 & 0.3017 & 0.4189 & 0.5112 & 0.5843 & Delta \\
 &  & \textbf{\cite{keshishzadeh2016improved}} & 0.8254 & 0.7935 & 0.7019 & 0.6368 & 0.6621 & 0.8018 & 0.8254 & Delta \\
 & \multirow{-3}{*}{\textbf{\begin{tabular}[c]{@{}l@{}}State-of\\ -the-art\end{tabular}}} & \textbf{\cite{gui2014exploring}} & 0.8711 & 0.8531 & 0.7556 & 0.6882 & 0.5101 & 0.7819 & 0.8711 & Delta \\ \cline{2-11} 
 & \textbf{} & \textbf{Att-RNN} & 0.9384 & 0.7928 & 0.8318 & 0.6854 & 0.6046 & 0.9238 & 0.9384 & Delta \\
 & \textbf{} & \textbf{Ours} & \textbf{1.0} & \textbf{0.9285} & \textbf{0.8366} & 0.5529 & 0.4558 & 0.9417 & 1.0 & Delta \\ \hline \hline
\multirow{13}{*}{\textbf{EID-M}} & \multirow{5}{*}{\textbf{\begin{tabular}[c]{@{}l@{}}Non-DL \\ Baseline\end{tabular}}} & \textbf{SVM} & 0.7796 & 0.5424 & 0.5664 & 0.6522 & 0.4915 & 0.7477 & 0.7796 & Delta \\
 &  & \textbf{RF} & 0.8124 & 0.7194 & 0.7351 & \textbf{0.6842} & 0.4765 & 0.8121 & 0.7194 & Delta \\
 &  & \textbf{KNN} & 0.8211 & 0.7501 & 0.7649 & 0.6611 & 0.3821 & 0.8162 & 0.7501 & Delta \\
 &  & \textbf{AB} & 0.3228 & 0.3095 & 0.2478 & 0.2548 & 0.2529 & 0.3189 & 0.3228 & Delta \\
 &  & \textbf{LDA} & 0.3082 & 0.1681 & 0.1621 & 0.1824 & 0.1311 & 0.2995 & 0.3082 & Delta \\ \cline{2-11} 
 & \multirow{3}{*}{\textbf{\begin{tabular}[c]{@{}l@{}}DL \\ Baseline\end{tabular}}} & \textbf{LSTM} & 0.8482 & 0.6926 & 0.7438 & 0.5726 & 0.5008 & 0.8185 & 0.8482 & Delta \\
 &  & \textbf{GRU} & 0.862 & 0.6935 & 0.7531 & 0.5672 & \textbf{0.5072} & 0.8221 & 0.862 & Delta \\
 &  & \textbf{CNN} & 0.7647 & 0.6712 & 0.7191 & 0.5588 & 0.4949 & 0.7749 & 0.7749 & Full \\ \cline{2-11} 
 & \multirow{3}{*}{\textbf{\begin{tabular}[c]{@{}l@{}}State-of\\ -the-art\end{tabular}}} & \textbf{\cite{jayarathne2016brainid}} & 0.9721 & 0.7019 & 0.7091 & 0.4189 & 0.4089 & 0.8195 & 0.9721 & Delta \\
&  & \textbf{\cite{keshishzadeh2016improved}} & 0.9931 & 0.6891 & 0.6988 & 0.5124 & 0.3397 & 0.7963 & 0.9931 & Delta \\
 &  &\textbf{\cite{gui2014exploring}} & 0.9917 & 0.7199 & 0.6572 & 0.4911 & 0.3977 & 0.8011 & 0.9917 & Delta \\ \cline{2-11} 
 & \textbf{} & \textbf{Att-RNN} & 0.9324 & 0.6847 & 0.6846 & 0.5732 & 0.4941 & 0.7976 & 0.9324 & Delta \\
 &  & \textbf{Ours} & \textbf{0.9996} & \textbf{0.9013} & \textbf{0.8989} & 0.4428 & 0.3661 & \textbf{0.8858} & 0.9996 & Delta\\ \hline \hline
\end{tabular}
}
\end{table*}

\begin{figure*}[t]
\centering
\begin{minipage}[b]{0.48\linewidth}
\centering
\includegraphics[width=\textwidth]{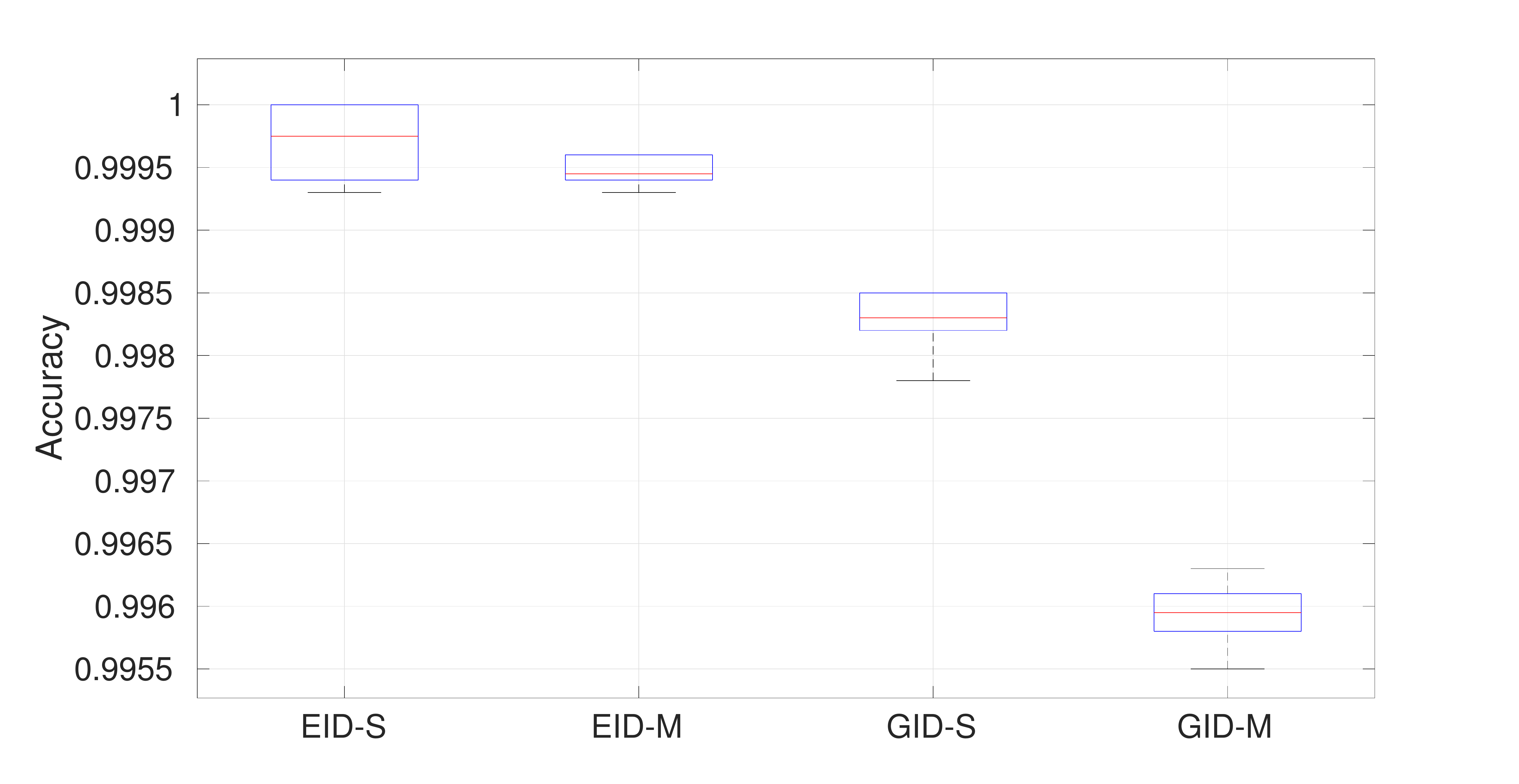}
\caption{Impact of Session}
\label{fig:sessions}
\end{minipage}
\begin{minipage}[b]{0.48\textwidth}
\includegraphics[width=\textwidth]{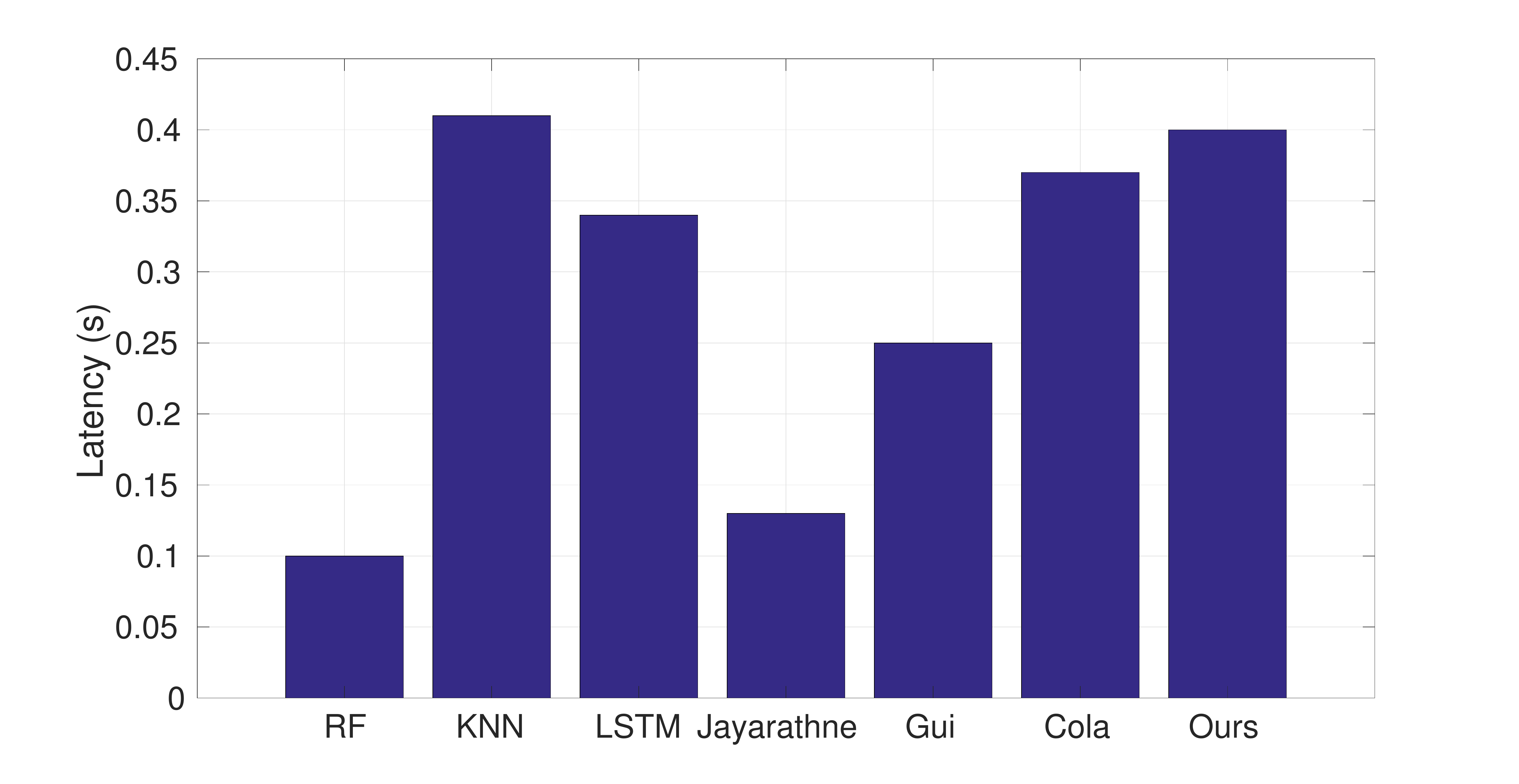}
\caption{Latency}
\label{fig:latency}
\end{minipage}
\end{figure*}

The primary conclusions from Table~\ref{tab:component_level_comparison} are summarized as follows: 
\begin{itemize}
  \item Our approach achieves the highest performance over both EEG and gait datasets under single session and multi-session settings;
  \item For most of the baselines, we observe lower accuracy in EEG dataset compared to that in gait dataset, implying the EEG-based authentication is still challenging and our approach has room to improve; nevertheless, our model outperforms others and show superiority to both EEG-based or Gait-based methods;
  \item The results on single-session datasets are better than that on multi-session datasets. This is reasonable, and it demonstrates that the number of sessions does affect the authentication accuracy. This problem will be further analyzed in Section~\ref{sub:impact_of_sessions};
  \item Our model achieves better performance than Att-RNN. Since the diversity between our model and Att-RNN is that we employ an external classifier, this observation demonstrates that the external classifier is effective and efficient.
\end{itemize}

In Deepkey, the subjects passing the Invalid ID Filter are regarded as genuine only if their recognized IDs are consistent, i.e., $E_{ID}=G_{ID}$. It can be inferred easily that the FAR of Deepkey is
0 as well. However, the FRR depends on one or more of these
three scenarios: the false rejection of Invalid ID Filter; the incorrect Gait identification; the incorrect EEG identification. In summary, the overall FAR is \textbf{0} and the overall FRR is calculated as $\mathbf{1\%}=(0.006+0.994*((1-0.9961)+0.9961*(1-0.9996)))$.

\subsection{Impact of Key Parameters} 
\label{sub:impact_of_key_parameters}

\subsubsection{Impact of sessions} 
\label{sub:impact_of_sessions}
In practical applications, sessions in different scenarios may result in a minor difference in equipment position, signal quality, and other factors. To investigate the impact of sessions, we conduct external experiments by comparing the performance between single-session datasets and multi-session datasets. The comprehensive evaluation metrics and the comparison over various baselines are listed in Table~\ref{tab:classification_report} and Table~\ref{tab:component_level_comparison} while the comparison is summarized in Figure~\ref{fig:sessions}. The experiment results show that on the multi-session datasets, compared the single-session datasets, we achieve a slightly lower but still highly competitive performance.

\subsubsection{Impact of EEG Band} 
\label{sub:impact_of_eeg_band}
A series of comparison experiments are designed to explore the optimal EEG frequency band which contains the most discriminative features. The results presented in Table~\ref{tab:bands_comparison} illustrate that:
\begin{itemize}
   \item The Delta band consistently provides higher identification accuracy compared to other frequency bands for both single and multiple sessions. This observation shows that Delta pattern contains the most discriminative information for person identification.
   \item Our method gains the best outcome on both datasets with different sessions. This validates the robustness and adaptability of the proposed approach.
 \end{itemize} 
Why Delta pattern could outperform other patterns since Delta wave mainly appear in deep sleep state? Here we give one possible reason. We know the fact that the EEG patterns are associated with individuals' mental and physical states (organics and systems). For example, while the subject is under deep sleep and producing Delta pattern, the most parts of physical functions of the body (such as sensing, thinking, even dreaming) are completely detached. Only the very essential life-support organs and systems (such as breathing, heart beating, and digesting) keep working, which indicates the brain areas corresponding to life-support functions are active. While the subject is awake (e.g., relax state) and producing Alpha pattern, the subject has more activated functions such as imaging, visualizing and concentrating. Also, more brain functions like hearing, touching, and thinking are attached, which means that more physical brain areas (such as frontal lobe, temporal lobe, and parietal lobe) are activated. At this time, the life-support organs are still working. In short, only the life-support organs related brain areas are active in the first scenario (Delta pattern) whilst the brain areas controlling life-support and high-level functions (e.g., concentrating) are active in the second scenario (Alpha pattern). Thus, we infer that the delta pattern is correspond to the life-support organs and systems, which is the most stable function in different scenarios and the most discriminative signal in inter-subject classification.

\subsubsection{Impact of Datasize} 
\label{sub:impact_of_datasize}
\begin{figure}
\centering
\includegraphics[width=0.5\textwidth]{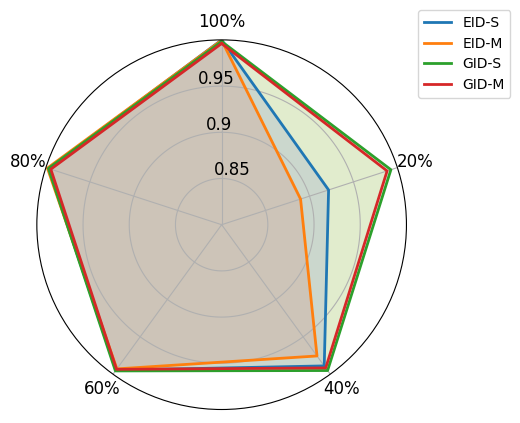}
\caption{Impact of Datasize. The spokes of this radar chart represent the datasize proportion (20\%, 40\%, 60\%, 80\%, 100\%) while the concentric circles indicate the classification accuracy (85\%, 90\%, 95\%, and 100\%).}
\label{fig:datasize}
\end{figure}

Datasize is one crucial influence element in deep learning based algorithms. In this section, we conduct experiments to train the proposed method over various data sizes. As shown in the radar chart (Figure~\ref{fig:datasize}), four datasets are evaluated under different proportion of training datasize. It has five  equi-angular spokes which represent the proportion of datasize (20\%, 40\%, 60\%, 80\%, 100\%), respectively. The four concentric circles indicate the accuracy which are 85\%, 90\%, 95\%, and 100\%, respectively. Each closed line represent a dataset and have five intersections with the five spokes. Each intersection node represents the classification accuracy of a specific dataset with a specific proportion datasize. For example, the intersection of EID-S line and 20\% spoke is about 0.92, which denote that our approach achieves the accuracy of 92\% over EID-S dataset with 20\% datasize.
The radar chart infers that gait datasets (GID-S and GID-M) obtain competitive accuracy even with 20 percent datasize; nevertheless, EID-S and EID-M highly rely on the datasize. This phenomenon is reasonable because EEG data has lower signal-to-noise ratio and requires more samples to learn the latent distribution pattern. 

\subsection{Latency Analysis} 
\label{sub:latency_analysis}
We also study the latency of Deepkey since low delay is highly desirable in real-world deployment.
The latency of Deepkey is compared to the response time of several state-of-the-art authentication approaches. The comparison results are shown in Figure~\ref{fig:latency}.
The latency of our method is 0.39 sec, which is competitive compared 
to the state-of-the-art baselines. 
Specifically, the reaction time of Deepkey is composed of three components, with the Invalid ID Filter taking 0.06 sec and the ID Identification taking 0.33 sec. 


\section{Discussions and Future Work} 
\label{sec:discussions_and_future_work}
In this paper, we propose a biometric identification system based both EEG and gait information. In this section, we discuss the open challenges and potential future research directions.

First, the datasets used in this paper only has limited participants. The extensive evaluations over more subjects are necessary. However, compare to some existing works (\cite{zhang2018mindid}, \cite{bashar2016human}, \cite{sun2014gait} have 8, 9, 10 subjects in their experiments, respectively), we believe our participants amount is acceptable.
Our work has already demonstrate that Deepkey can be used in settings such as small offices which are accessed by a small group of people.
In addition, evaluation can be improved by extending observations on how the system performs in different conditions. For example, considering changes in EEG signals during more trails, and longer times (hours, days or even months) to understand if these are consistent and reliable for detection.

Second, wearable sensors like EEG headset and wearable IMUs are required in the data collection stage of Deepkey system. Extensive experiments are meaningful in the future to investigate how the placement position of the wearable sensor matters. The EEG headset position on the head and the IMU position on the arms may affect the authentication performance. With the development of hardware related techniques, the EEG headset is becoming more portable and affordable. For example, the developed cEEGrids\footnote{http://ceegrid.com/home/}, an flex-printed, multi-channel sensor arrays that are placed around the ear using an adhesive, are easy and comfortable to wear and dispatch. This a good tendency of EEG acquisition equipment although the cEEGrids are currently expensive. 
In the future, the widely deployment of DeepKey authentication system in real world environment is possible. 

Moreover, another future scope is to develop device-free authentication techniques to overstep the inconvenient brought by wearable sensors. We have analyzed the computational latency in Section~\ref{sub:latency_analysis}. However, the overall system latency not only include the computation latency but also include the data collection latency such as the time cost when wear the EEG headset and the IMUs. The data collection latency of our system is about ten seconds while EEG collection requires about four seconds and gait collection requires above six seconds. Compare to other authentication system such as fingerprint, our data collection latency is much higher. This is one of the major future directions of our work. A potential solution is develop device-free authentication system and measure gait signals by environmental sensors such as RFID (Radio Frequency IDentification) tags.
The Received Signal Strength Indication (RSSI) of RFID tag measures the present received signal power which reflects the target subject's walking information.

Additionally, a promising future work is the real-world online deployment of the proposed DeepKey system. Since we have demonstrated that the effectiveness and efficiency of Deepkey in off-line situation, the future step is to build a real-time authentication environment to evaluate the online performance.

Furthermore, the proposed DeepKey still facing the challenge of `in the wild' scenario since the gathered EEG data are easily corrupted by physical actions like walking. In this work, the EEG and gait data are gathered in two separate steps. However, in the outdoor environments, the user is hardly stand still and wait for the authentication. Fortunately, our system has competitive performance in a fixed indoor environment (such as bank vouchers) which can provide a stable data collection environment and mainly concerned about the high fake-resistance.

\section{Conclusion} 
\label{sec:conclusion}
Taking the advantages of both EEG- and gait-based systems for fake resistance, we propose Deepkey, a multimodal biometric authentication system, to overcome the limitations of traditional unimodal biometric authentication systems. 
This authentication system Deepkey contains three independent models: an Invalid ID Filter Model, a Delta band based EEG Identification Model, and a Gait Identification Model, to detect invalid EEG data, recognize the EEG ID and Gait ID, respectively. 
The Deepkey system outperforms the state-of-the-art baselines by achieving a FAR of $0$ and a FRR of $1\%$. In addition, the key parameters (such as sessions and EEG frequency band) and the system latency are also investigated by extensive experiments.

This work sheds the light on further research on multimodal biometric authentication systems based on EEG and gait data. Our future work will focus on deploying the Deepkey system in an online real-world environment. In addition, the gait signals are currently gathered by three wearable IMUs, which may obstruct the large-scale deployment in
practice. Therefore, another direction in the future is to collect gait data from non-wearable gait solutions (e.g., sensors deployed in environments).

\bibliographystyle{ACM-Reference-Format}
\bibliography{deepkey.bib}

\end{document}